\documentclass{article}
\usepackage[utf8]{inputenc}
\usepackage{graphicx}
\usepackage{caption}
\usepackage{subcaption}
\usepackage{xcolor}
\usepackage{booktabs}
\usepackage{multirow}
\usepackage{amsmath}
\usepackage{setspace}
\usepackage{placeins}
\usepackage{authblk}

\begin{document}

\title{Revisiting Agglomerative Clustering}
\author[a]{Eric K. Tokuda}
\author[b]{Cesar H. Comin}
\author[a]{Luciano da F. Costa}
\affil[a]{S\~ao Carlos Institute of Physics, University of S\~ao Paulo, S\~ao Carlos, SP, Brazil}
\affil[b]{Department of Computer Science, Federal University of S\~ao Carlos, S\~ao Carlos, SP, Brazil}

\maketitle

\begin{abstract}
An important issue in clustering concerns the avoidance of false positives while searching for clusters. This work addressed this problem considering agglomerative methods, namely single, average, median, complete, centroid and Ward's approaches applied to unimodal and bimodal datasets obeying uniform, gaussian, exponential and power-law distributions. A model of clusters was also adopted, involving a higher density nucleus surrounded by a transition, followed by outliers. This paved the way to defining an objective means for identifying the clusters from dendrograms. The adopted model also allowed the relevance of the clusters to be quantified in terms of the height of their subtrees. The obtained results include the verification that many methods detect two clusters in unimodal data. The single-linkage method was found to be more resilient to false positives. Also, several methods detected clusters not corresponding directly to the nucleus. The possibility of identifying the type of distribution was also investigated.
\end{abstract}

\section{Introduction}

The task of categorization, also called classification, can take place according to the two following main paradigms: (i) \emph{supervised classification}, in which prototypes and samples of pre-defined categories are available and used for comparison with the entities to be classified; and (ii) \emph{unsupervised classification}, or \emph{clustering}, characterized by the absence of any previous knowledge about the number of properties of the categories, which have to be otherwise abstracted~\cite{duda2012pattern}. Needless to say, the latter type of classification is more challenging, being also particularly important as it presupposes any supervised scheme.  

Clustering is not an easy endeavour, as it depends on several factors such as the choice of the features to be considered as well as on the type of classification method to be adopted.  Indeed, especially when the entities belonging to different categories are not markedly distinct, different choices of features and/or clustering methods can lead to substantially diverging results.  These challenges have motivated the development of several clustering approaches, of which hierarchical methods have received special attention.  Hierarchical clustering can proceed by successively \emph{dividing} or \emph{agglomerating} the original entities into groups and subgroups according to some criteria~\cite{jain1999data}.  As such, hierarchical methods provide valuable information about the interrelationship between categories.  

Hierarchical clustering methods may construct the hierarchy in two opposite directions, bottom-up  (agglomerative) and top-down (divisive). Despite similar in concept, they may eventually come up with different solutions~\cite{kaufman2009finding}. Divisive algorithms are more robust in the early stages compared to its counterpart~\cite{macnaughton1964dissimilarity} and agglomerative clustering techniques, in turn, are more understandable and by far the most popular~\cite{jain1999data}.

The choice of the agglomerating criterion defines the respective hierarchical clustering method.  These criteria are typically based on distances between clusters, such as the single-linkage approach, which considers the smallest distance between the points in two clusters.  However, it is also possible to consider the minimization of dispersion of the points inside each cluster as a criterion, which gives rise to methods such as Ward's agglomerative clustering~\cite{ward1963hierarchical}.  In the present work, we consider the following classical methods:  single-linkage, complete-linkage, centroid, median, average, and Ward's.

The coexistence of many methods for the same finality often motivates efforts aimed at identifying their respective relative performance concerning specific size and types of data, choice of features, number of categories, etc.  Therefore, it is not surprising that agglomerative hierarchical clustering methods have been studied comparatively (e.g.~\cite{roux2018comparative,reynolds2006clustering}).  Often, these methods are investigated concerning their potential to separate existing clusters.  

While the identification of existing clusters is undoubtedly an important performance factor to be taken into account, other aspects should be considered.   Of particular relevance concerns the resilience to false-positive identification of clusters.  This is often critical because data devoid of clusters should not lead to any significant cluster. 

The comparative evaluation of the performance of the adopted agglomerative hierarchical methods regarding their robustness to false-positive constitutes the main objective of the present work.  In order to do so, we consider several types of clusterless point distributions -- more specifically uniform, linear, exponential, normal, and power-law -- as input to the adopted agglomerative hierarchical methods.   A relevance parameter is also proposed, capable of quantifying the prominence of the detected clusters, which is used to investigate the robustness of the adopted methods when applied to the above indicated clusterless datasets.  More specifically, it would be expected that these methods never, or rarely, detect two or more clusters.

This article is organized as follows.  In Section~\ref{sec:related} we expose the most relevant works in the literature, in  Section~\ref{sec:method} some basics concepts and the cluster identification methodology are presented. The experiments are shown and discussed in Section~\ref{sec:experiments}. In Section~\ref{sec:conclusion} we present the final considerations.

\section{A Brief Review of Agglomerative Clustering}
\label{sec:related}

The hierarchical clustering studies date back to the 50s, with Florek~\cite{florek1951liaison} and McQuitty~\cite{cohen2013research} independently proposing the foundations of the nearest-neighbour (single linkage) cluster analysis method. Since then, a variety of new forms of hierarchical clustering has been proposed as well as applied with different ends~\cite{dubes1976clustering}.  In \cite{massart1983non}, the authors propose an exact and a heuristic clustering methods. They also proposed a general \emph{robustness analysis} of each method to select the most relevant clusters. Motivated by these robustness analysis, the authors of \cite{plastria1986two} performed the robustness analysis for different numbers of clusters and a subsequent step to obtain the data hierarchy.

Most hierarchical cluster analysis consider a symmetric dissimilarity. In \cite{ozawa1983classic}, the authors observed that symmetric measurements may not be appropriate for every data set and proposed an algorithm based on \emph{asymmetric} similarities. The approach was validated with gestalt clusters. 
Real data may present additional challenges, such as errors and missing data. The method proposed by~\cite{lee2005hierarchical} handle noisy data by optimizing the ordinal consistency between the similarity data and the hierarchical structure. It also efficiently deals with incomplete data by using the available similarity data available.

Inspired by dynamical systems and quantum mechanics, \cite{zhang2013agglomerative} proposed an agglomerative clustering procedure in which clusters are characterized by the path integral and the affinity between clusters by incremental path integral.  They considered absorbing random walks and achieve state-of-art results with synthetic and imagery data. 
In~\cite{lu2013pha}, the authors extended the concept of potential field to clustering. The potential field produced by all points is used in conjunction with the dissimilarity matrix to obtain a balance between global and local information. The algorithm additionally utilizes efficient data structures to achieve reduced time complexity.

Dasgupta~\cite{dasgupta2016cost} frames the hierarchical clustering problem as a combinatorial optimization problem, with the cost given by the similarity between the points. The definition of the problem from the optimization standpoint allows the computation of complexity and the comparison of different algorithms. In this optimization framework, dissimilar elements are at higher levels of hierarchy and when the similarities between elements are the same, they achieve the same cost. This optimization task is NP-hard, but it has a provably good  approximation ratio. These ideas inspired subsequent authors, such as \cite{cohen2019hierarchical} to analyze multiple practical hierarchical clustering algorithms.

Hybrid methods, based on density and hierarchical information, have gained special attention from the pattern recognition community.  In~\cite{cheng2019hierarchical}, the authors focus on noisy data and propose a two-stage approach: noise is initially removed based on the density of the points and the remaining points are sequentially merged resulting in a hierarchical structure. In~\cite{campello2013density}, the authors start by building a graph from the points using mutual reachability; then a hierarchy of connected components is obtained by traversing the graph in increasing order of distance; the branches with low number of points are considered as not being associated to any cluster. This approach is also aimed at detecting outliers.

\section{Methodology}
\label{sec:method}

\subsection{Linkage methods}

Given a dataset containing $N$ elements, where the $i$-th element is represented by a feature vector $\vec{x}_i$, the initial step of agglomerative clustering is to link the points together to form candidate clusters according to a given metric. This is known as the \emph{linkage} step.

Let the distance between any two clusters $w$ and $v$ be represented as $D(w,v)$. Many distinct metrics can be used for defining $D$, but here we only consider Euclidean distances. Initially, each cluster contains a single object, that is, the number of clusters is given by the number of objects. At each iteration, pairs of clusters that were not previously joined are inspected, and the pair of clusters having the minimum value in $D$ is joined to form a new cluster $u$. Let $s$ and $t$ be the pair of clusters that were joined, and define $x[i]$ to represent an object $i$ in cluster $x$. After the clusters are joined, the distances $D(u,v)$ between the new cluster $u$ and all other clusters need to be calculated. Many methods have been defined to calculate the new distances, and they have a large impact on the resulting hierarchy. Common approaches are described below.

Single-linkage (SIN)~\cite{florek1951liaison,sneath1957application,mcquitty1957elementary}: the distance between clusters $u$ and $v$ is given by the smallest distance between every pair of objects $(u[i], v[j])$, that is,

\begin{equation}
    D(u,v) = \min_{ij}\left\{ D(u[i], v[j])\right\}
\end{equation}

Complete-linkage (COM)~\cite{sorensen1948method}: calculated as the largest distance between every pair of objects between clusters $u[i]$ and $v[j]$,

\begin{equation}
    D(u,v) = \max_{ij}\left\{ D(u[i], v[j])\right\}
\end{equation}

Average-linkage (AVG)~\cite{sokal1958statistical}: also called Unweighted Pair Group Method with Arithmetic Mean (UPGMA)~\cite{kaufman2009finding}, defines the new distance as the average distance between objects in $u$ and $v$, that is,

\begin{equation}
    D(u,v) = \sum_{ij}\frac{D(u[i], v[j])}{|u||v|},
\end{equation}
where $|x|$ represents the number of objects in cluster $x$.

Centroid-linkage (CEN)~\cite{sokal1958statistical}: also called Unweighted Pair Group Method with Centroid (UPGMC)~\cite{kaufman2009finding}, uses the Euclidean distance between the clusters centroids,

\begin{equation}
    D(u,v) = ||C_u - C_v||,\label{eq:cen}
\end{equation}
where $C_x$ indicates the centroid of cluster $x$.

Median-linkage (MED)~\cite{gower1967comparison}: also called Weighted Pair Group Method with Centroid (WPGMC)~\cite{kaufman2009finding}, uses Equation~\ref{eq:cen} to calculate the distances, but the centroid of cluster $u$ is calculated as

\begin{equation}
    C_u = \frac{C_s + C_t}{2}.
\end{equation}

Ward-linkage (WAR)~\cite{ward1963hierarchical}: joins clusters leading to the minimum increase in within-cluster variance. This can be done by setting the distances as

\begin{equation}
    D(u,v) = \sqrt{\frac{|v|+|s|}{T}D(v,s)^2 + \frac{|v|+|t|}{T}D(v,t)^2 - \frac{|v|}{T}D(s,t)^2},
\end{equation}
where $T=|v|+|s|+|t|$.

The single-linkage method can be implemented using a minimum spanning tree~\cite{gower1969minimum}. The complete, average, weighted and ward methods are commonly implemented using the nearest-neighbor chain algorithm~\cite{mullner2011modern}. An efficient implementation for the centroid and median methods can be found in~\cite{mullner2011modern} and a detailed comparison of Ward's implementation is provided in~\cite{murtagh2014ward}.

\begin{figure}[t]
    \centering
    \begin{subfigure}[b]{0.3\textwidth}
        \includegraphics[width=\textwidth]{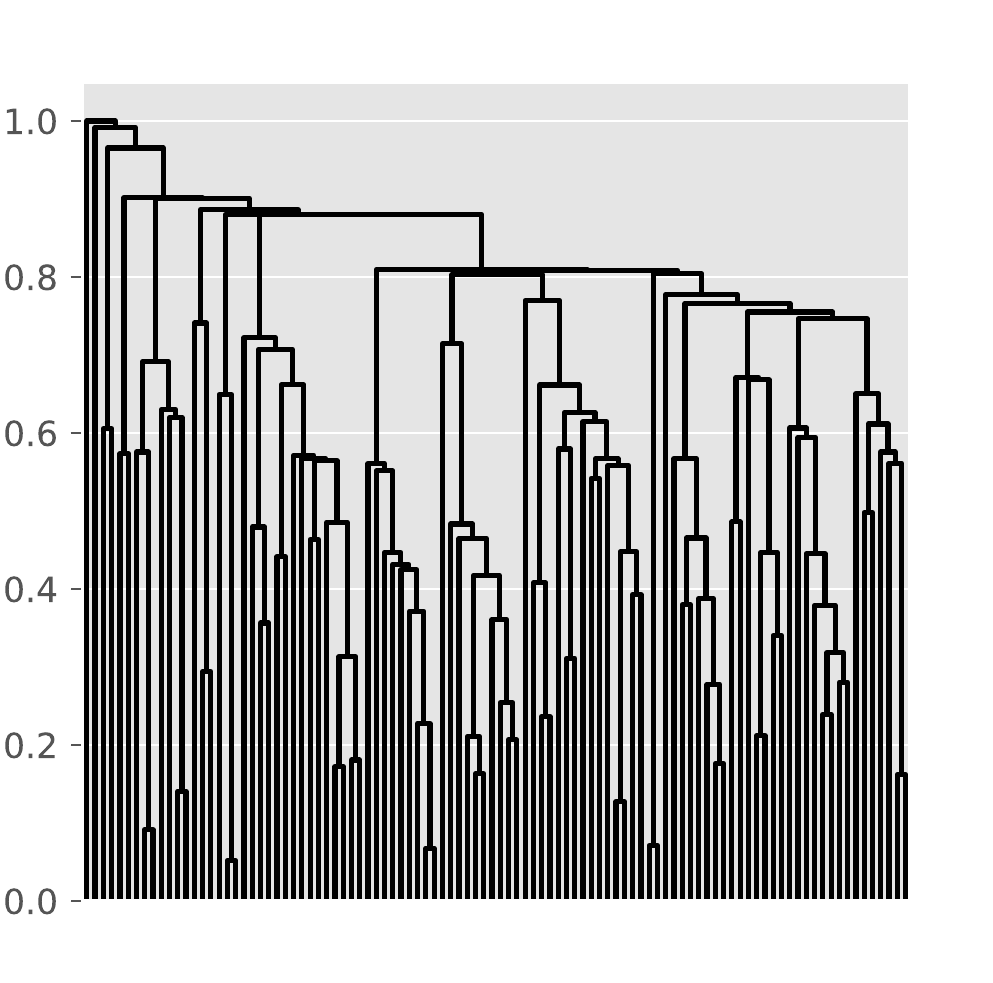}
        \caption{Single}
    \end{subfigure}
    \begin{subfigure}[b]{0.3\textwidth}
        \includegraphics[width=\textwidth]{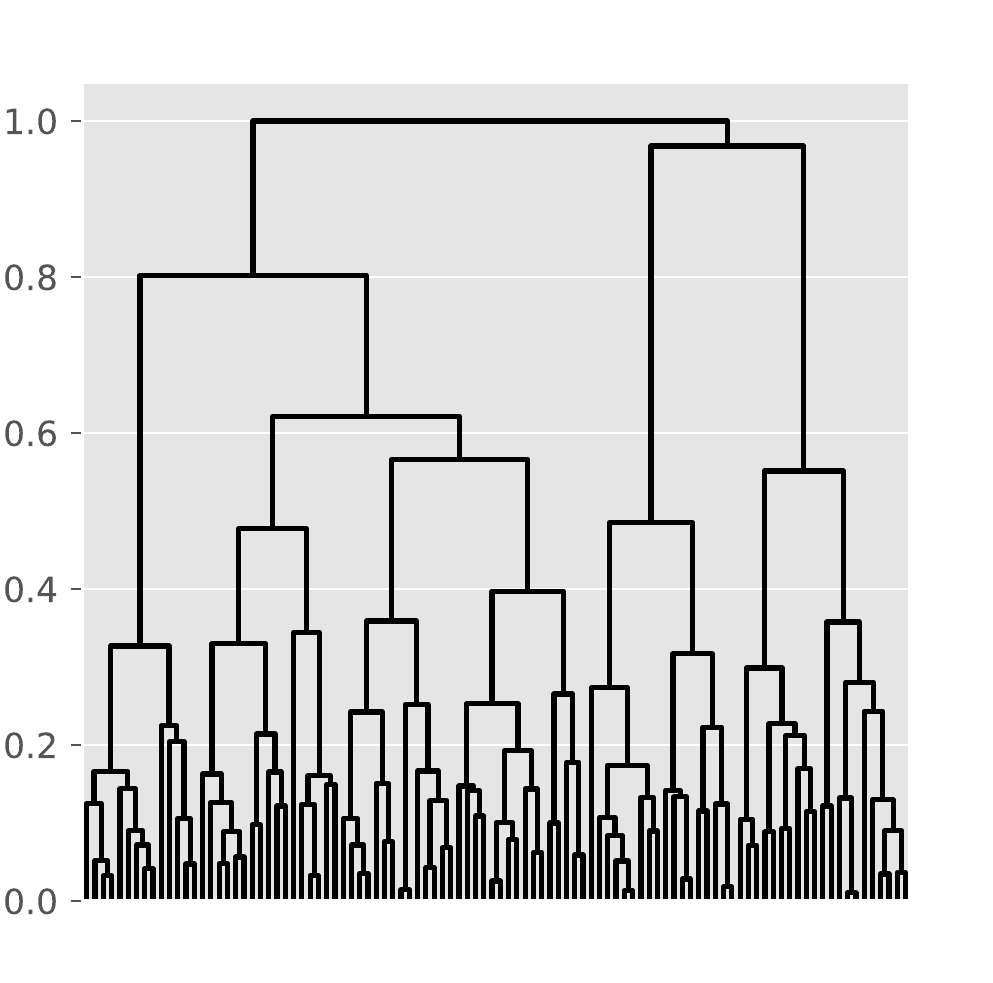}
        \caption{Average}
    \end{subfigure}
    \begin{subfigure}[b]{0.3\textwidth}
        \includegraphics[width=\textwidth]{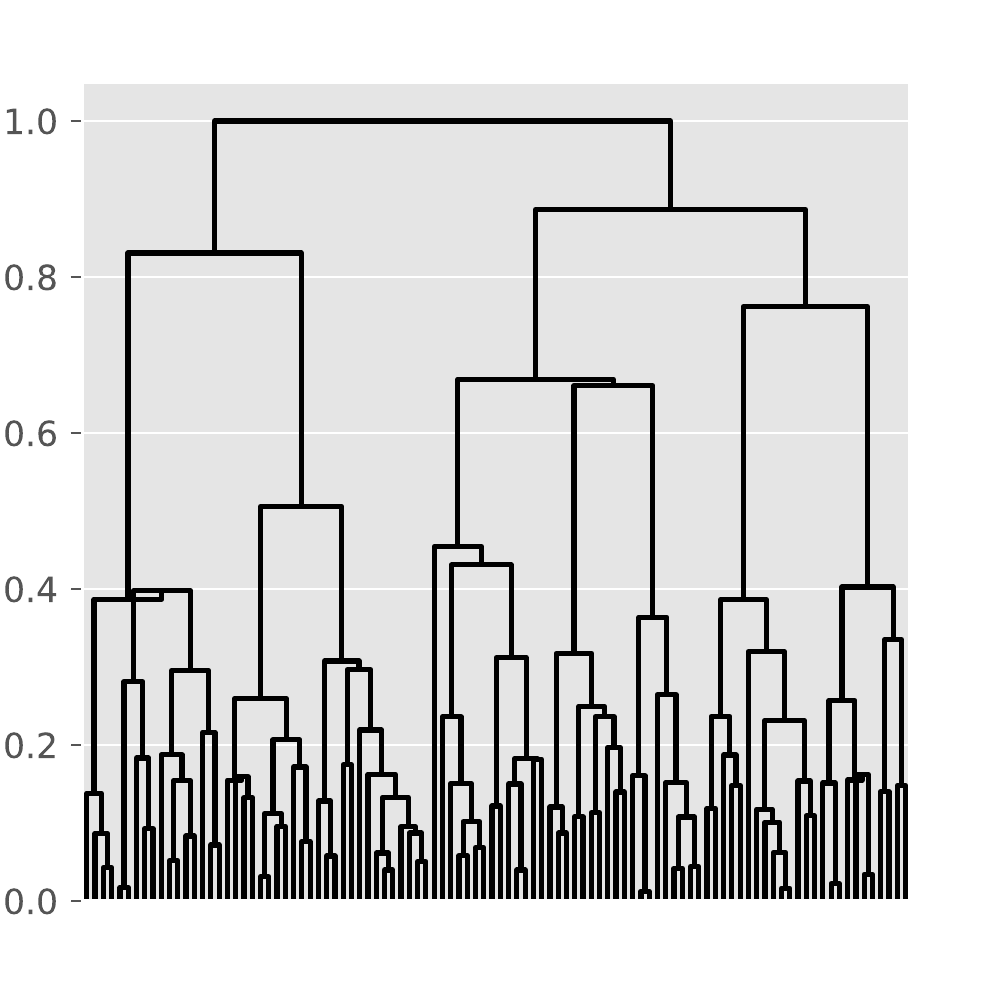}
        \caption{Median}
    \end{subfigure}\\
    \begin{subfigure}[b]{0.3\textwidth}
        \includegraphics[width=\textwidth]{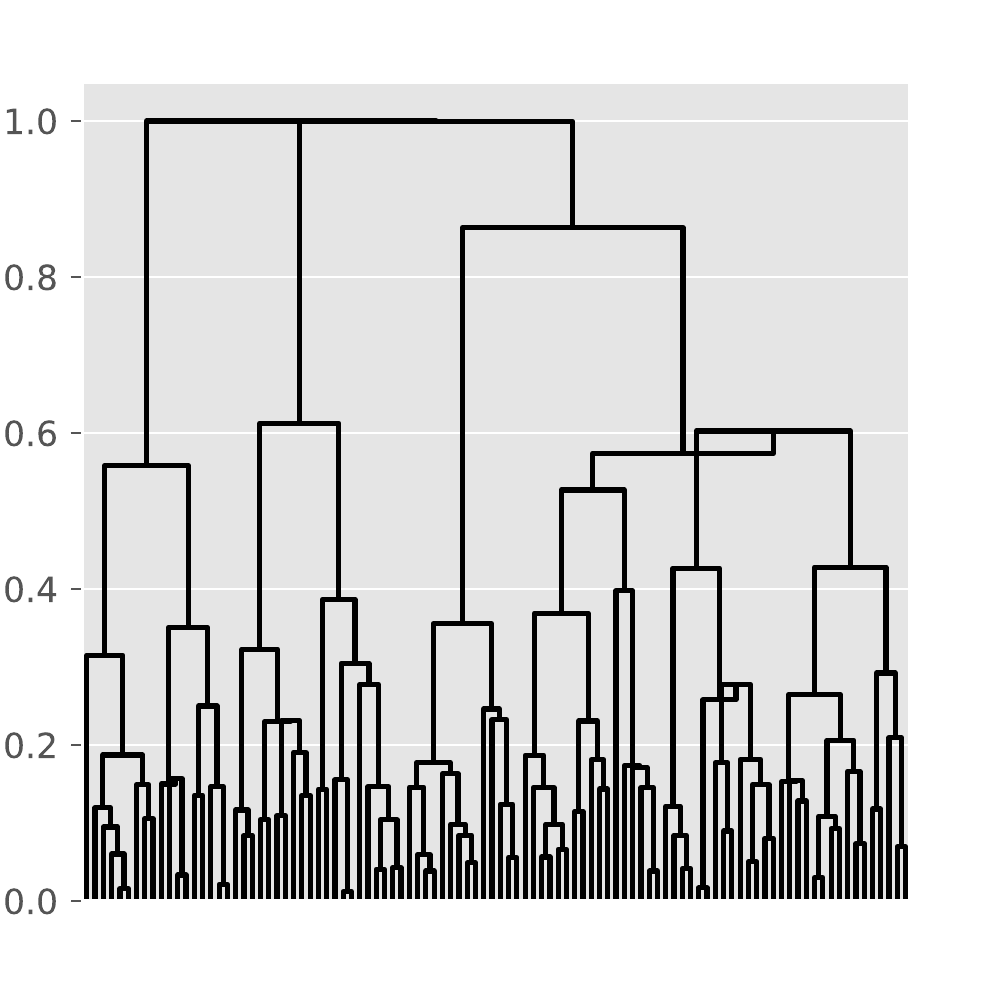}
        \caption{Centroid}
    \end{subfigure}
    \begin{subfigure}[b]{0.3\textwidth}
        \includegraphics[width=\textwidth]{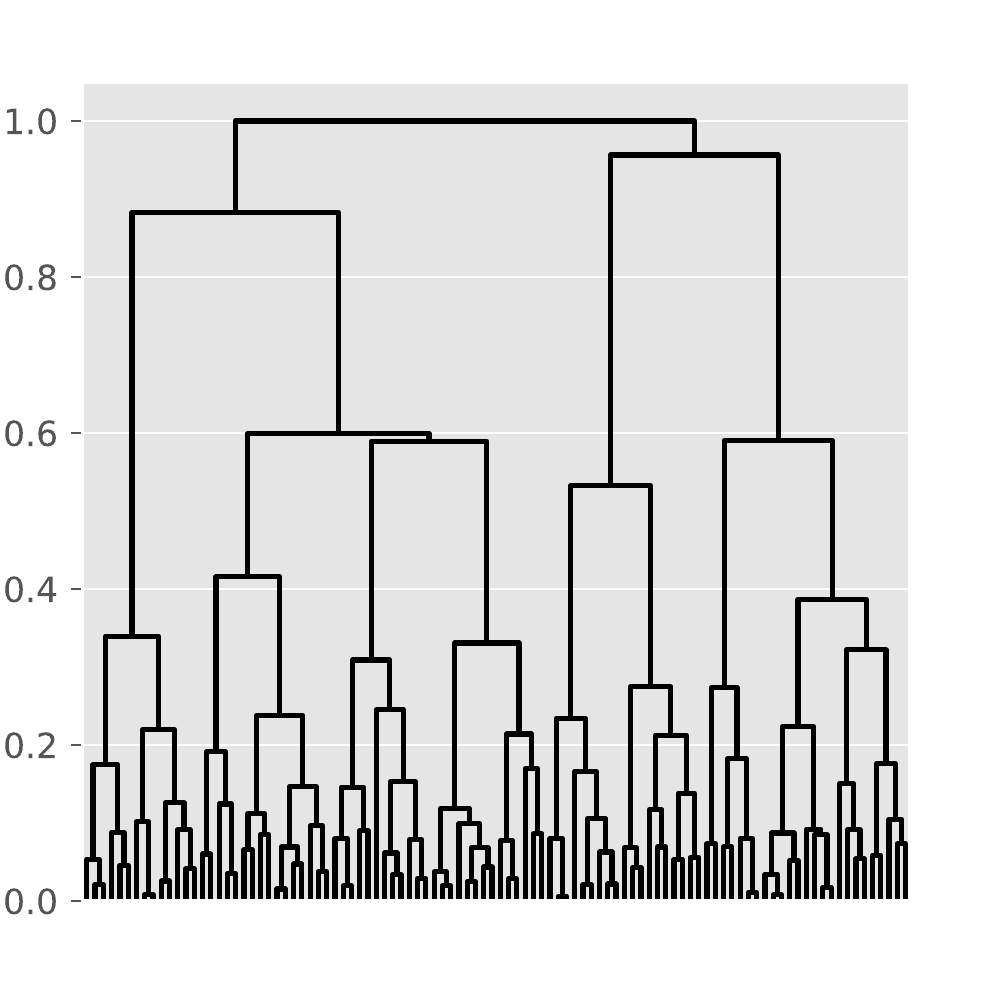}
        \caption{Complete}
    \end{subfigure}
    \begin{subfigure}[b]{0.3\textwidth}
        \includegraphics[width=\textwidth]{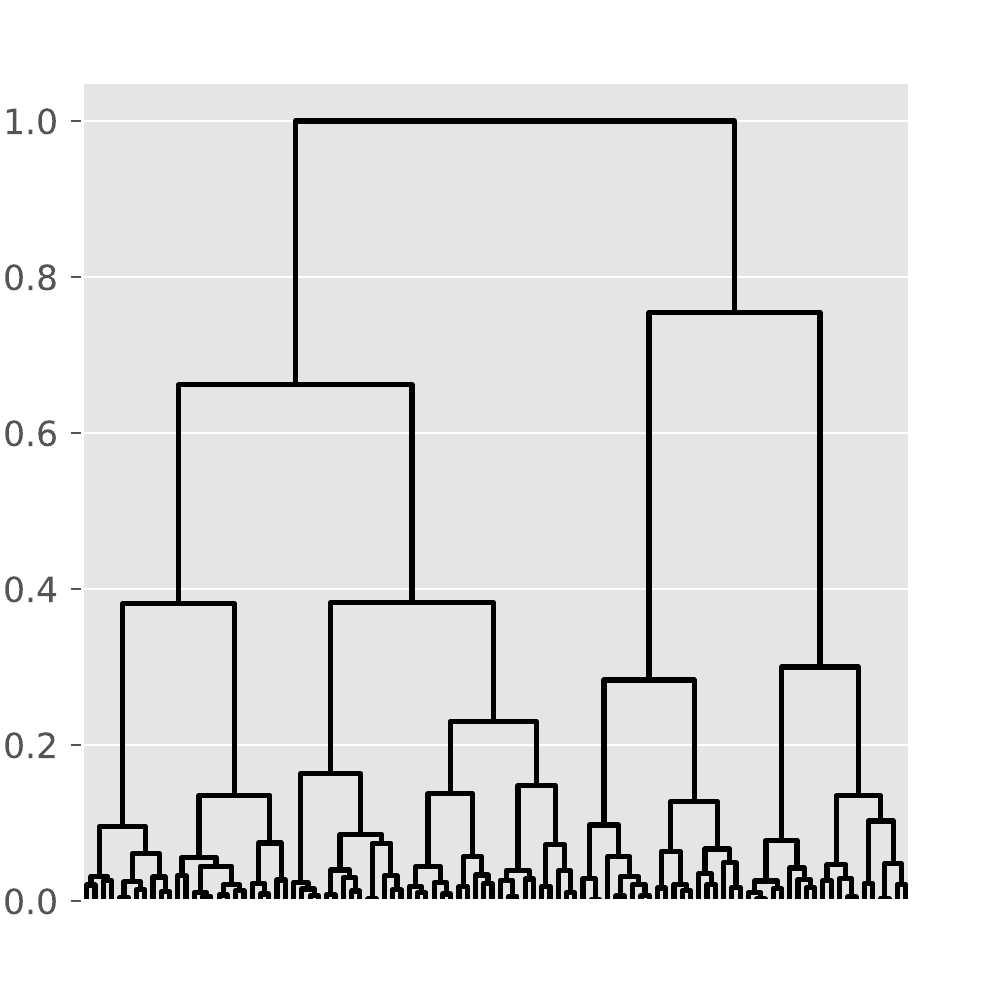}
        \caption{Ward}
    \end{subfigure}
    \caption{Dendrogram visualization of the hierarchical clusterization using a uniform randomly distributed set of points in a circle.}
    \label{fig:dendrograms}
\end{figure}

Figure~\ref{fig:dendrograms} shows the dendrograms obtained by the considered methods when applied to a uniform distribution of points inside a circle.

\subsection{Anatomy of a Cluster}
\label{subsec:anatomy}

One of the challenges in cluster detection is that the
selection, parameter setting or even the development of 
effective methodologies rely critically on the properties 
of the typically expected clusters (e.g.~\cite{jain1999data}).  
Hierarchical approaches are no exception to this, especially
given that the existing alternatives typically rely on
distinct agglomeration schemes (e.g.~minimal/maximal 
distance, dispersion, etc.).

\begin{figure}[ht!]
    \centering
    \begin{subfigure}[b]{0.32\textwidth}
        \includegraphics[width=\textwidth]{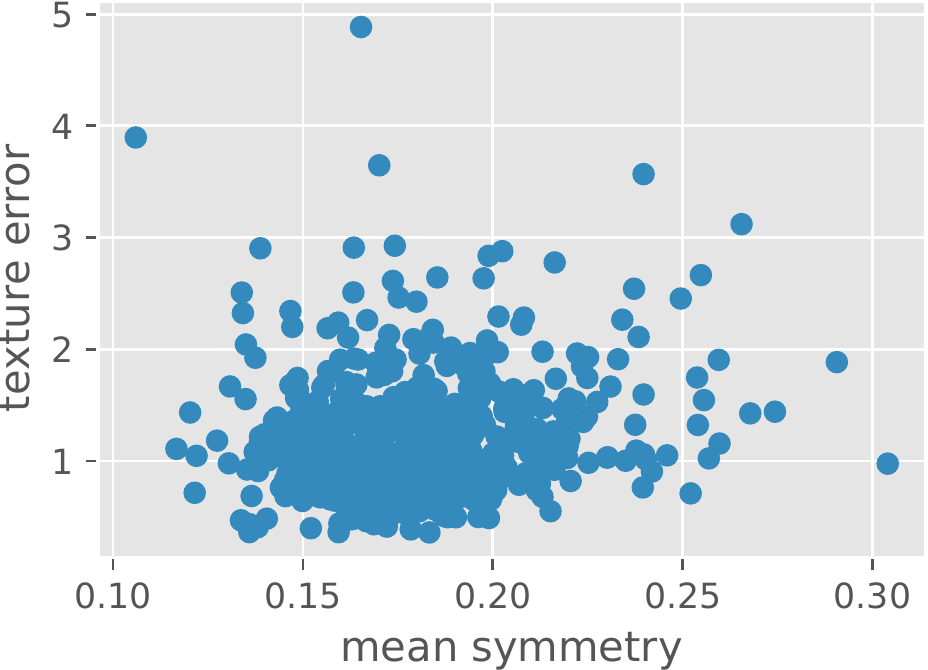}
        \caption{Cancer~\cite{street1993nuclear}}
    \end{subfigure}
    \begin{subfigure}[b]{0.32\textwidth}
        \includegraphics[width=\textwidth]{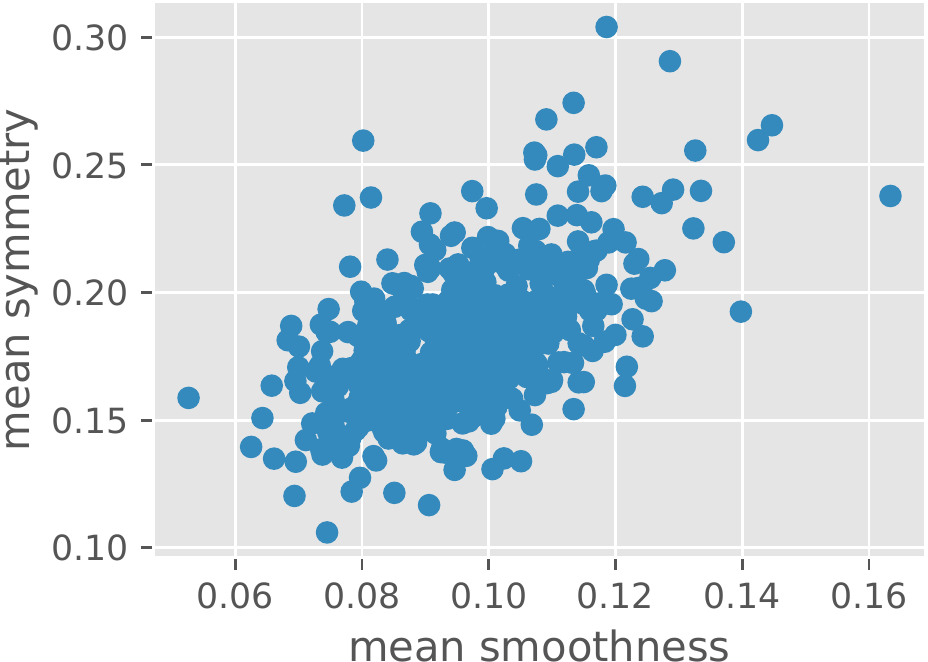}
        \caption{Cancer~\cite{street1993nuclear}}
    \end{subfigure}
    \begin{subfigure}[b]{0.32\textwidth}
        \includegraphics[width=\textwidth]{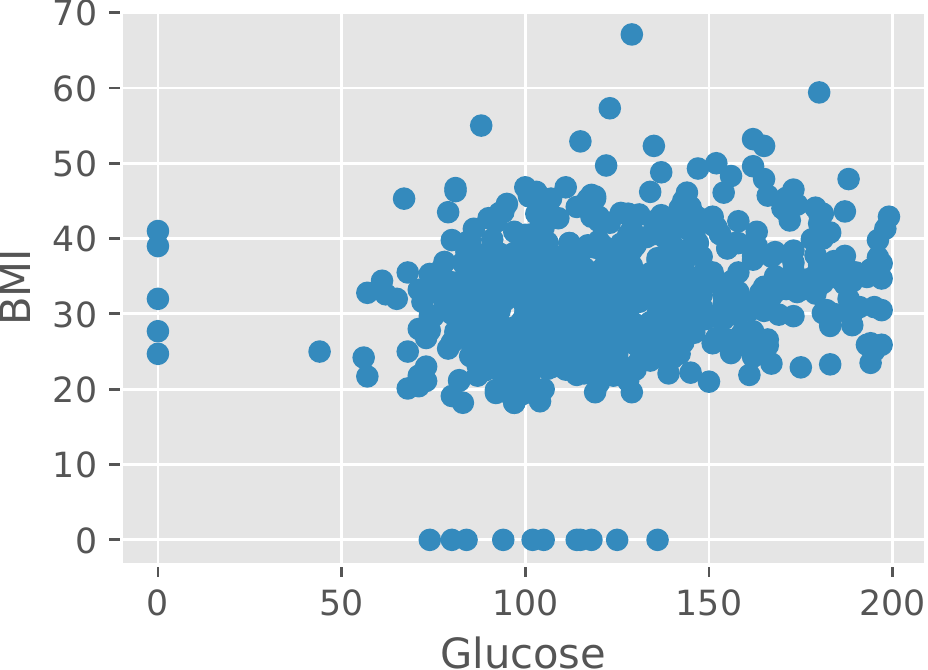}
        \caption{Diabetes~\cite{smith1988using}}
    \end{subfigure} \\
    \begin{subfigure}[b]{0.32\textwidth}
        \includegraphics[width=\textwidth]{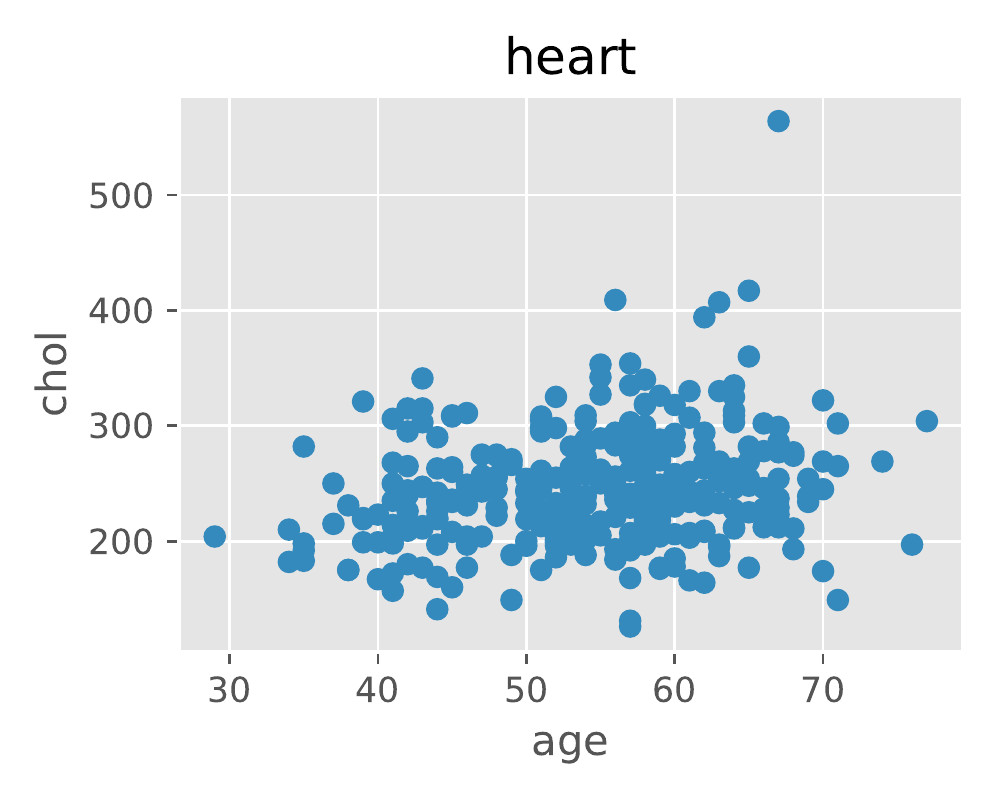}
        \caption{Heart disease~\cite{detrano1989international}}
    \end{subfigure}
    \begin{subfigure}[b]{0.32\textwidth}
        \includegraphics[width=\textwidth]{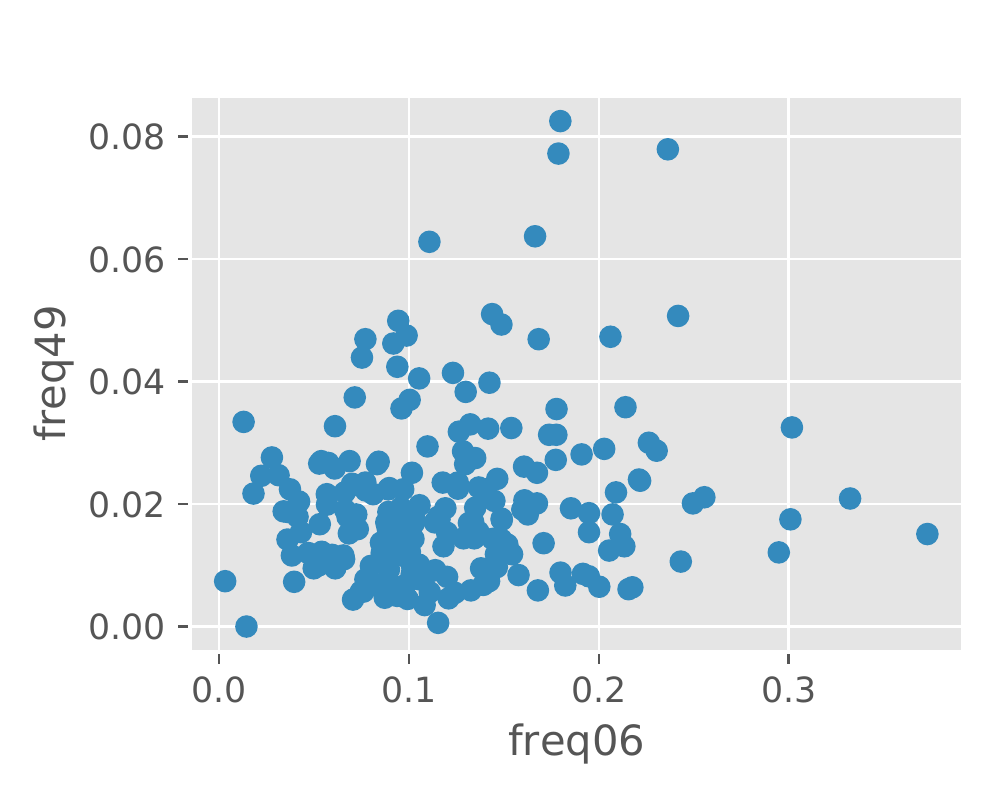}
        \caption{Sonar~\cite{gorman1988analysis}}
    \end{subfigure}
    \begin{subfigure}[b]{0.32\textwidth}
        \includegraphics[width=\textwidth]{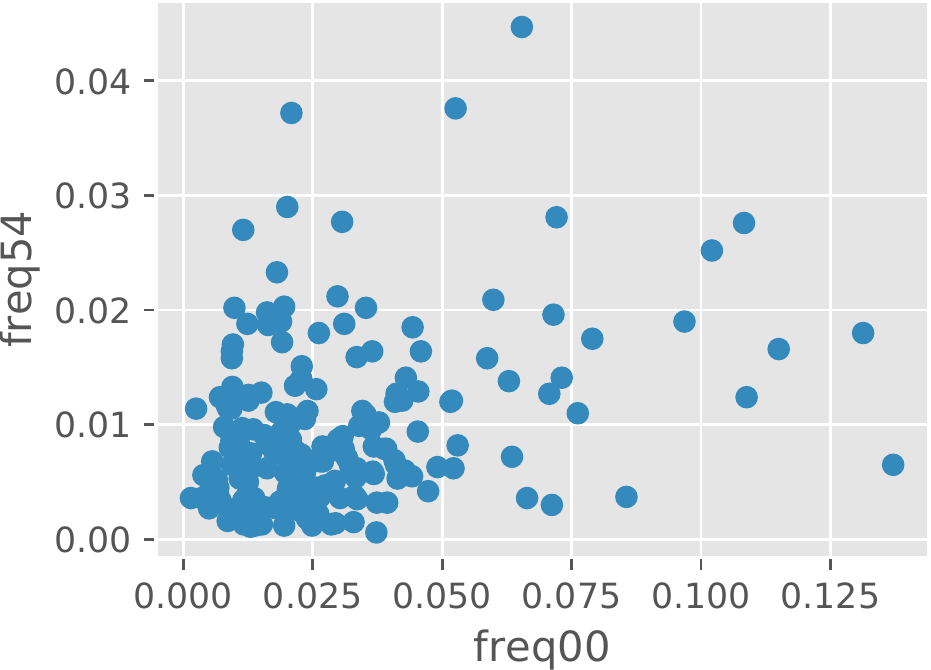}
        \caption{Sonar~\cite{gorman1988analysis}}
    \end{subfigure} \\
    \begin{subfigure}[b]{0.32\textwidth}
        \includegraphics[width=\textwidth]{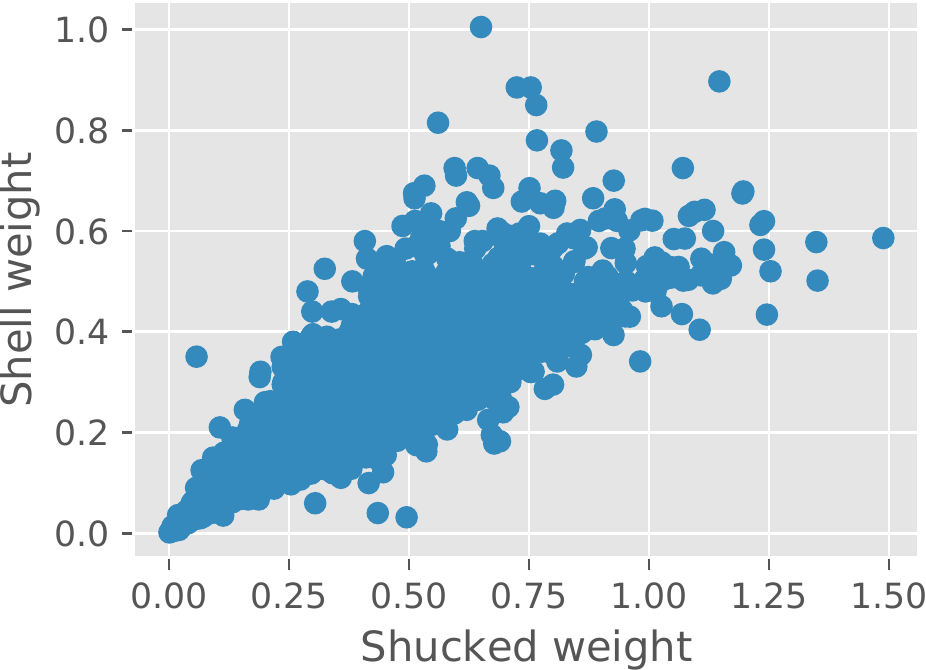}
        \caption{Abalone~\cite{warwick1994the}}
    \end{subfigure}
    \begin{subfigure}[b]{0.32\textwidth}
        \includegraphics[width=\textwidth]{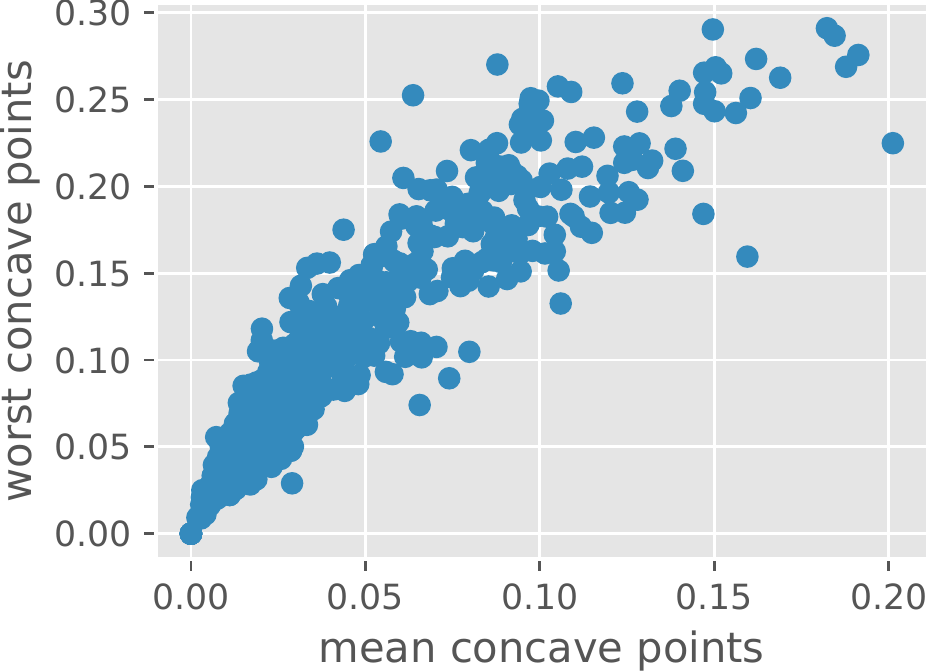}
        \caption{Cancer~\cite{street1993nuclear}}
    \end{subfigure}
    \begin{subfigure}[b]{0.32\textwidth}
        \includegraphics[width=\textwidth]{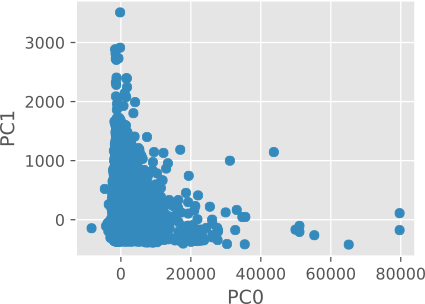}
        \caption{Bank (PCA)~\cite{moro2014data}}
    \end{subfigure} \\
        \caption{Scatterplots of pairs of features derived from problems of different nature (indicated in the respective titles) that present cluster structure compatible with that assumed in the present work, with the density of points increasing towards the respective nucleus, and the presence of a peripheral region and outliers.}
        \label{fig:pointsreal}
\end{figure}

Though a consensual, general definition of a cluster remains
elusive, several real-world data yield respective point 
distributions in feature spaces that exhibit a
a gradient of point density.
For instance, Figure~\ref{fig:pointsreal} depicts nine examples of real-data obtained from~\cite{uci}
mapped into a feature space that are characterized by this
type of point distributions. A previous work studying density organization of clusters has been reported in~\cite{campello2013density}.

\begin{figure}[t]
    \centering
    \begin{subfigure}[b]{0.24\textwidth}
        \includegraphics[width=\textwidth]{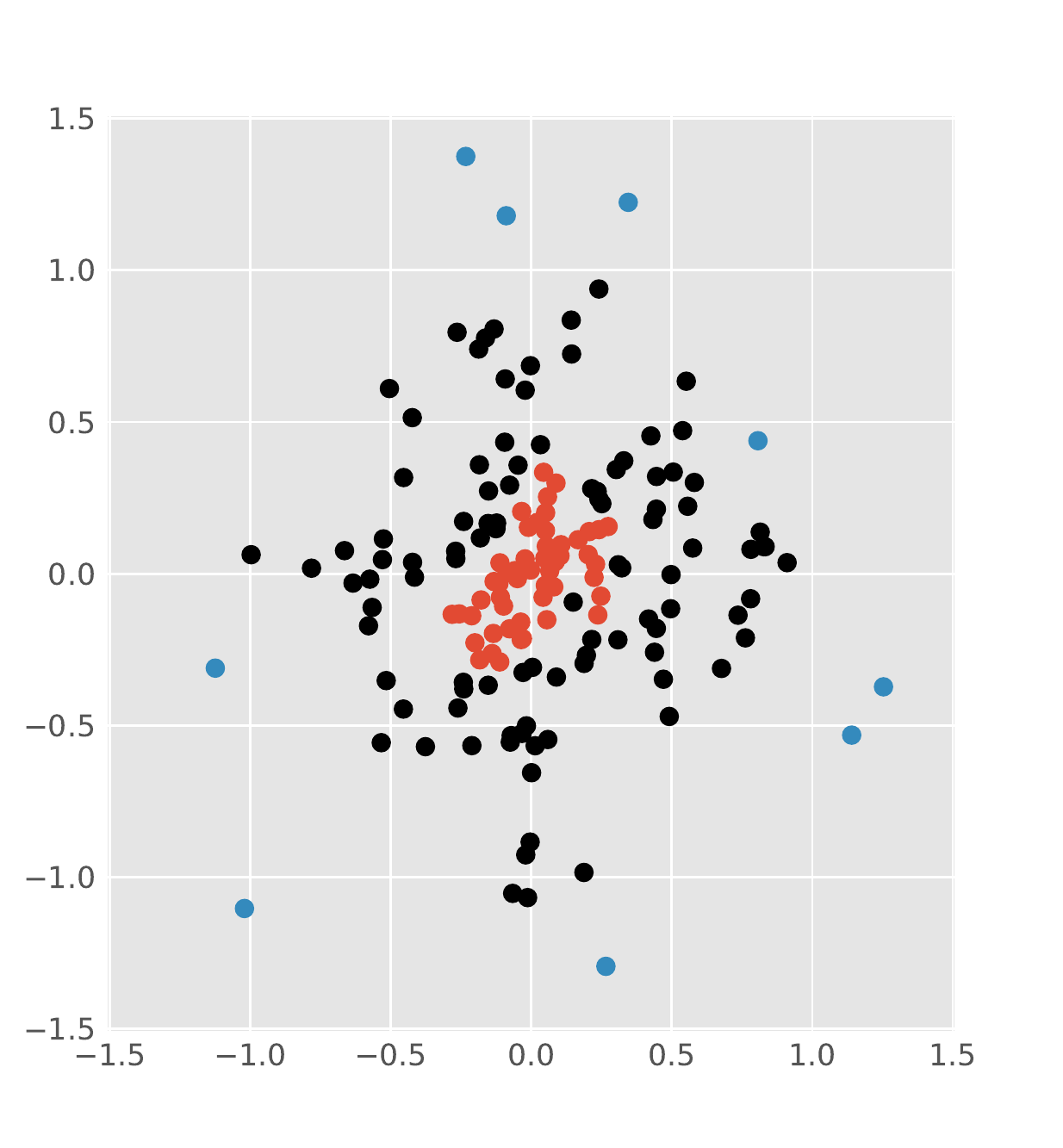}
    \end{subfigure}
    \begin{subfigure}[b]{0.24\textwidth}
        \includegraphics[width=\textwidth]{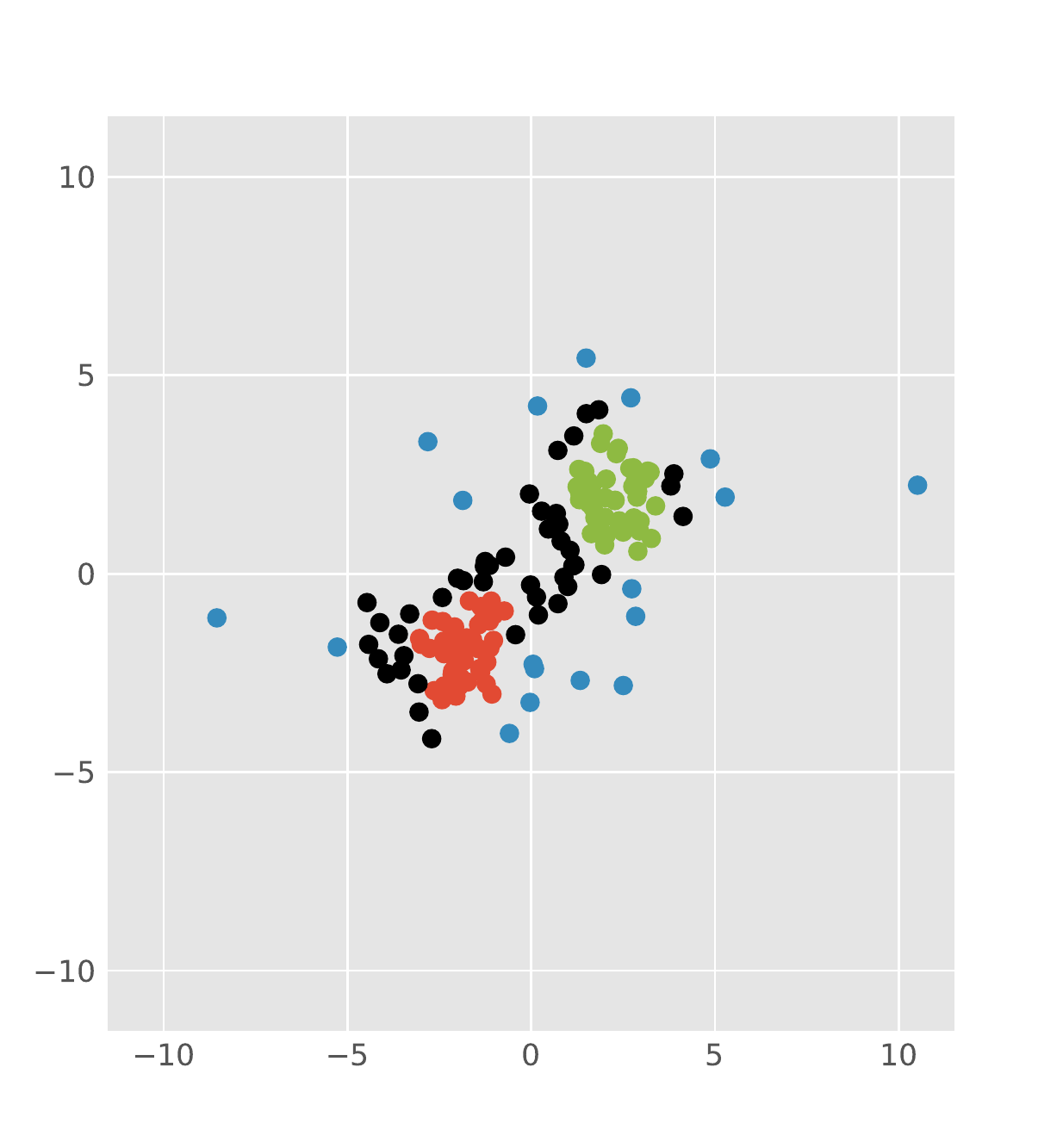}
    \end{subfigure}
    \begin{subfigure}[b]{0.24\textwidth}
        \includegraphics[width=\textwidth]{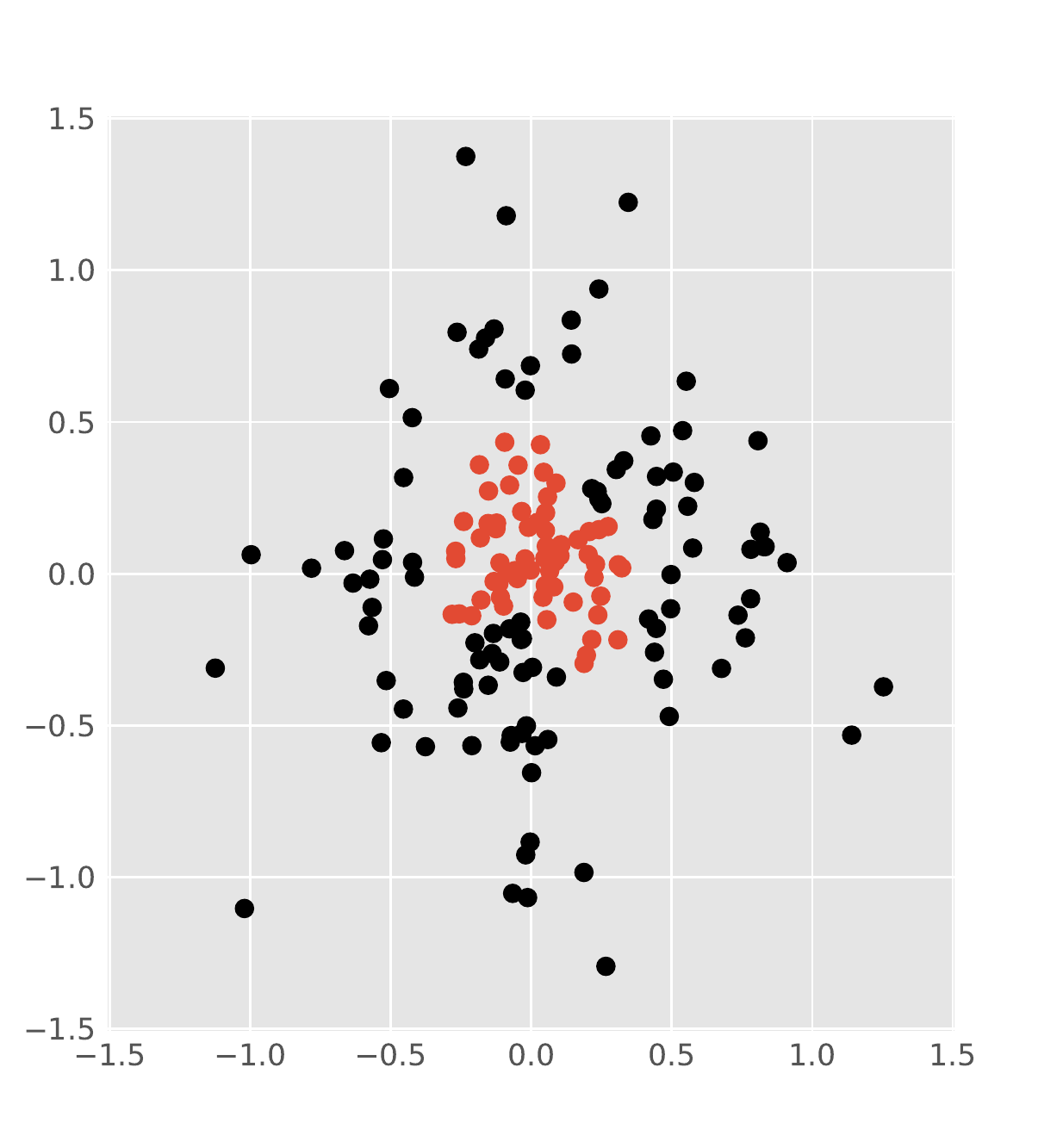}
    \end{subfigure}
    \begin{subfigure}[b]{0.24\textwidth}
        \includegraphics[width=\textwidth]{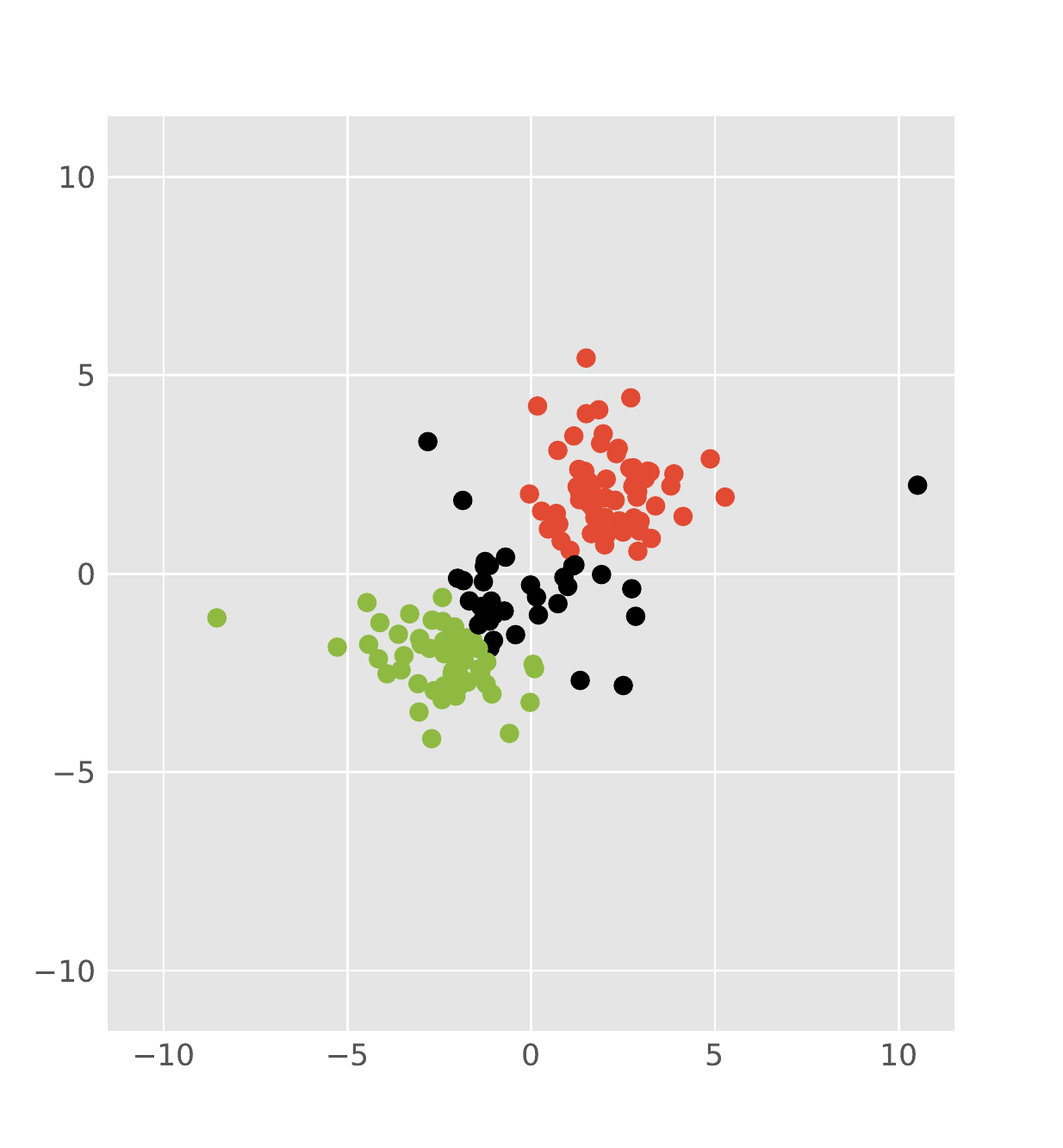}
    \end{subfigure} \\
    \begin{subfigure}[b]{0.24\textwidth}
        \includegraphics[width=\textwidth]{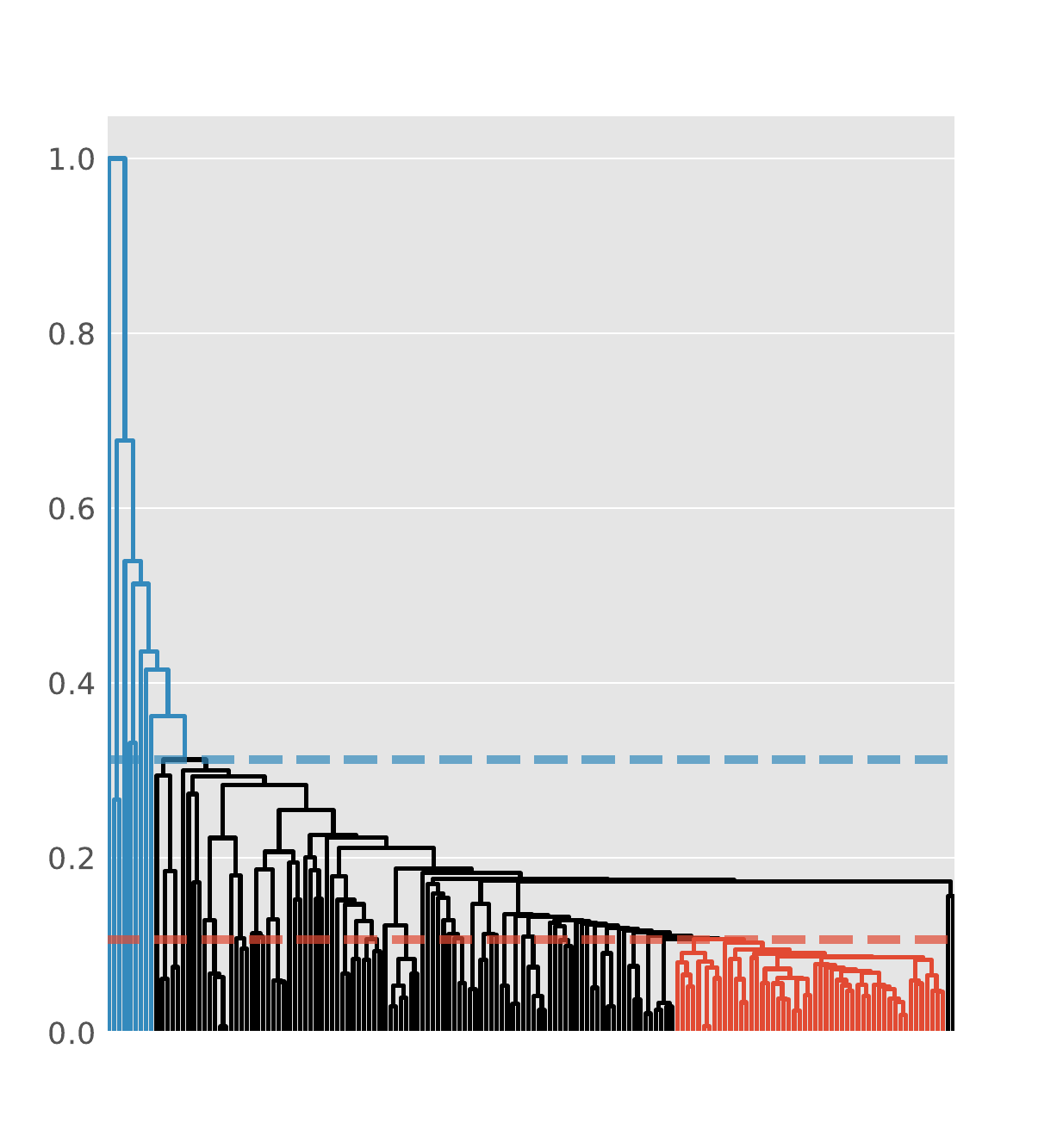}
        \caption{Unimodal Single}
    \end{subfigure}
    \begin{subfigure}[b]{0.24\textwidth}
        \includegraphics[width=\textwidth]{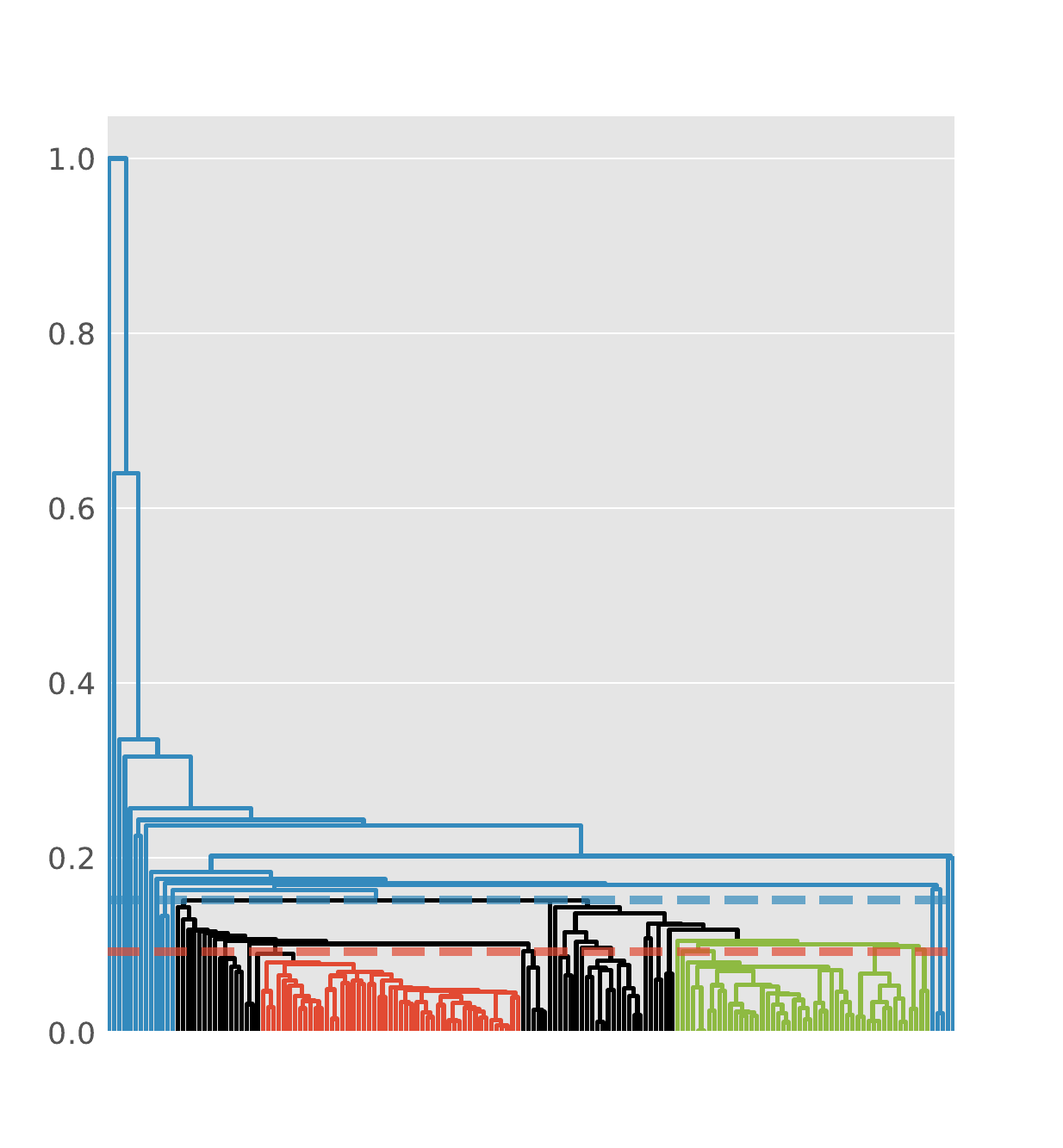}
        \caption{Bimodal Single}
    \end{subfigure}
    \begin{subfigure}[b]{0.24\textwidth}
        \includegraphics[width=\textwidth]{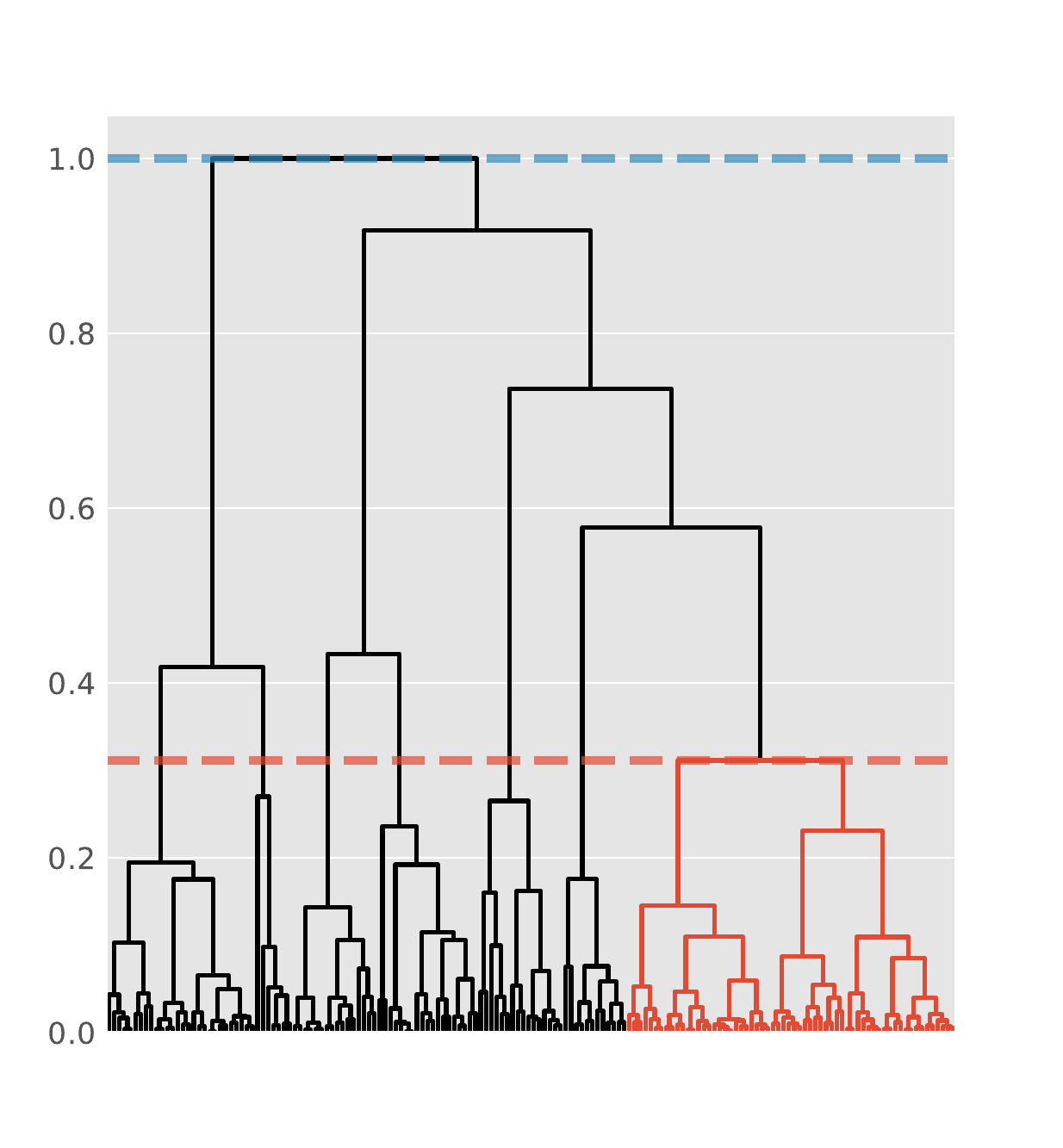}
        \caption{Unimodal Ward}
    \end{subfigure}
    \begin{subfigure}[b]{0.24\textwidth}
        \includegraphics[width=\textwidth]{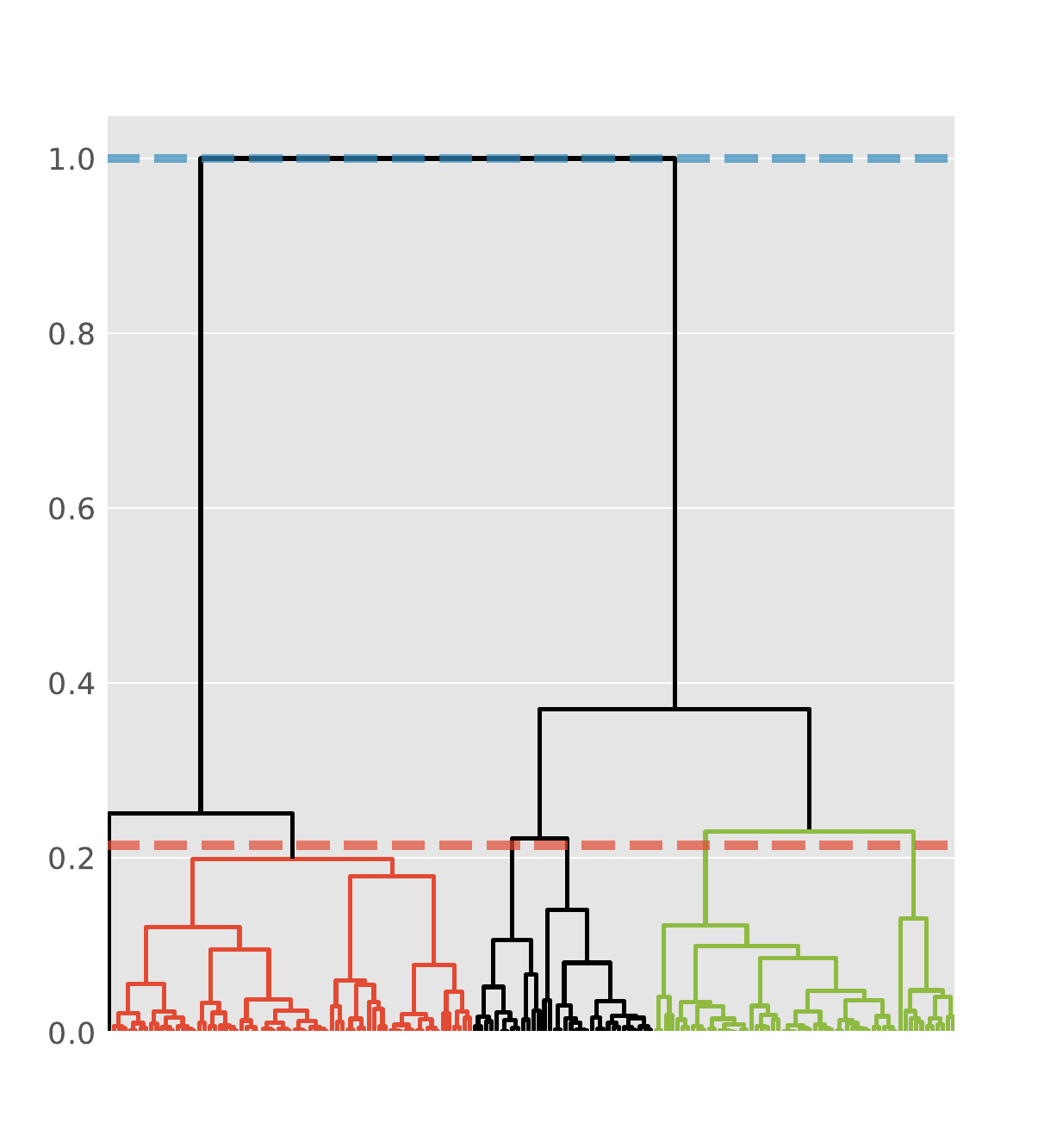}
        \caption{Bimodal Ward}
    \end{subfigure}
    
        \caption{Cluster identification using the Single and Ward linkage algorithms. Representation of the clusters (red and green), the transition (black) and the outlier (blue) regions in the dendrogram and the corresponding points in unimodal and in bimodal random exponentially distributed sets of points.}
        \label{fig:regionssingle}
\end{figure}

Figure~\ref{fig:regionssingle} 
illustrates the model of cluster considered
in the current work.  It is characterized by a high
density nucleus (red), followed by a medium density 
transition region (black), and then a peripheral low density 
region (blue).  Though the models shown in Figure~\ref{fig:regionssingle} have circular
symmetry, this is not a necessary condition for our approach.
The regions shown in this figure were identified by
the methodology to be described in Section~\ref{sec:method}.

Interestingly, these three regions tend to be well-preserved
even when more than one cluster is present, as illustrated
in Figures~\ref{fig:regionssingle}(b) and \ref{fig:regionssingle}(d).
Moreover, several important types of statistical point distribution,
such as those adopted in this work (uniform, Gaussian, power
and exponential) are all characterized by presenting the three 
proposed types of regions when including one or more clusters.
The adoption of a reference model as that shown in Figure~\ref{fig:regionssingle} allows several benefits.  

First, it allows the prediction of the behavior
of clustering approaches, which  need to be capable of treating
the varying density of points in an effective way.  In addition,
the availability of such models allow parameters of a chosen
method to be set more properly, in a semi-supervised fashion.
For instance, the number of points in the 3 regions --- nucleus, 
transition and periphery --- provides, as we shall see in this
work, an important subsidy not only for identifying clusters,
but also to assigning relevance levels to the obtained results.

Second, the peripheral, relatively isolated points, can be 
understood as  corresponding to \emph{outliers}, or it is also 
possible to conceive
methods that focus on the detection of the nucleus and then, as 
post-processing, tries to rescue the peripheral points in case
they are wanted.  However, it seems reasonable that the consideration
of very low-density regions simultaneously during cluster detection
represents an additional, potentially substantial challenge to 
the considered algorithms.

Third, the identification of properties of the three parts of 
clusters provides a subsidy for trying to
infer the type of point distributions characterizing the
clusters by using patter recognition resources.

The higher density of the nucleus provides more statistical
relevance, being potentially less affected by noise and
distortions.  Even more importantly, in multimodal datasets,
the nuclei of the existing clusters are almost invariably
 more separated one another than would be
the case if the other two regions were also taken into account.
This is in contrast to several existing methodologies, which
treat all density regions in a unified manner.

The varying density characterizing the adopted model has
another important implication, namely that it would be
interesting that the adopted clustering be capable of 
aggregating points into the nucleus in a progressive and relatively
smooth manner, to avoid instabilities as the cluster
merge along the obtained dendrograms.

Once the nuclei have been detected, hopefully with enhanced
quality, it is always possible to perform some post-processing
oriented to recovering the points in the transition and/or
periphery regions or each detected group.

\subsection{Cluster identification}

After the application of a linkage methodology to a dataset, the results are usually visualized using a dendrogram, such as those shown in Figure~\ref{fig:dendrograms}. Next, a methodology for selecting a suitable \emph{cut} of the dendrogram is applied, so as to define the clusters. For instance, a common strategy is to calculate the inconsistency~\cite{jain1988algorithms} of each merge performed during the linkage process, that is, each dendrogram bifurcation, and to find merges such that all the descendants have an inconsistency lower than a given threshold. Such a criterion usually disregard important prior knowledge one may have about the data, such as an estimation of the size of the clusters or the number of outliers expected in the data. An important situation is when the number of elements for each class is known, or can be estimated. In this case, one should search for clusters having specific sizes. Furthermore, the size of the clusters can also be predicted in cases where a prototype of the clusters is available or can be developed. 

\begin{figure}[t]
    \centering
    \begin{subfigure}[b]{0.3\textwidth}
        \includegraphics[width=\textwidth]{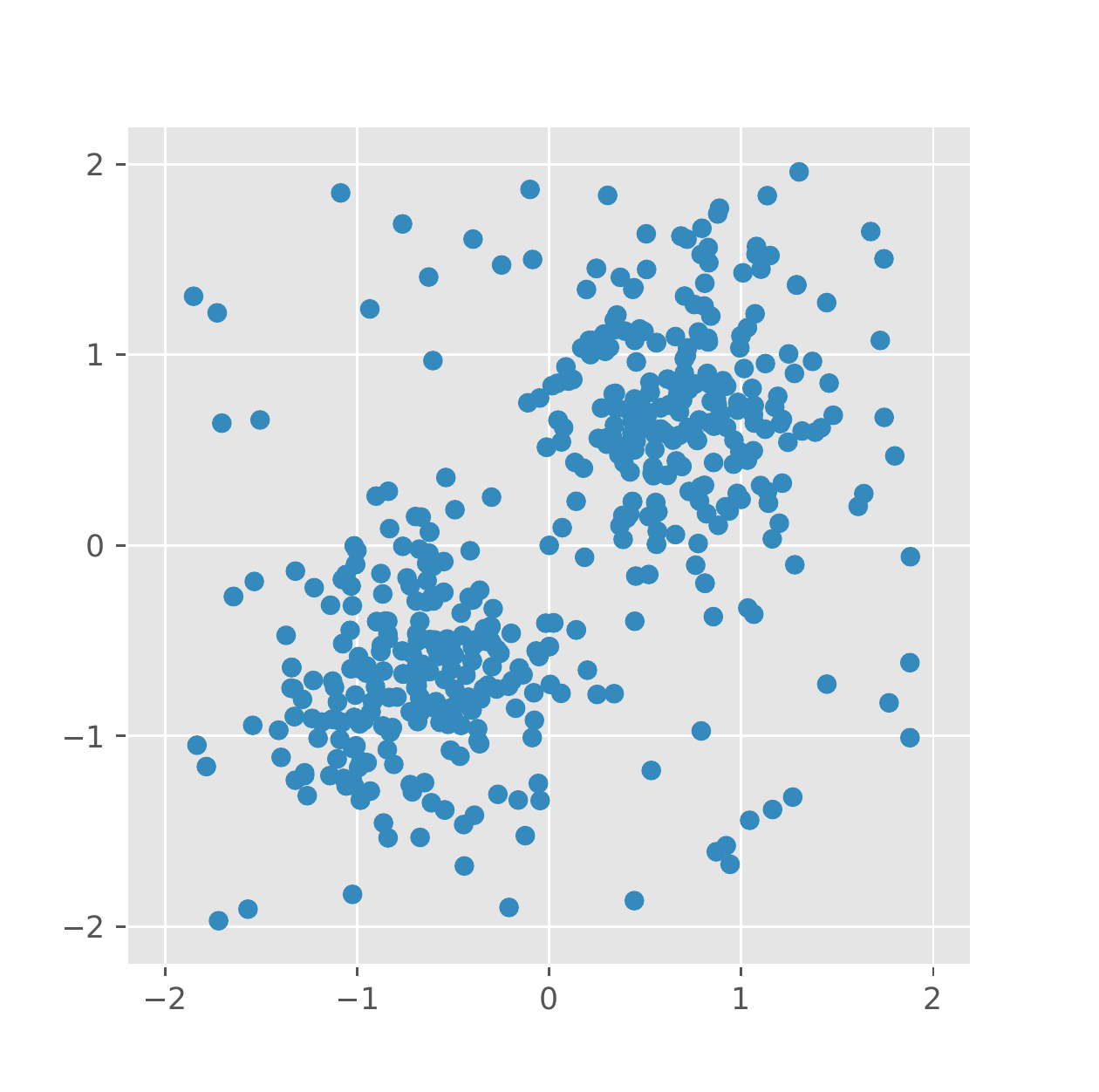}
    \end{subfigure}
    \begin{subfigure}[b]{0.3\textwidth}
        \includegraphics[width=\textwidth]{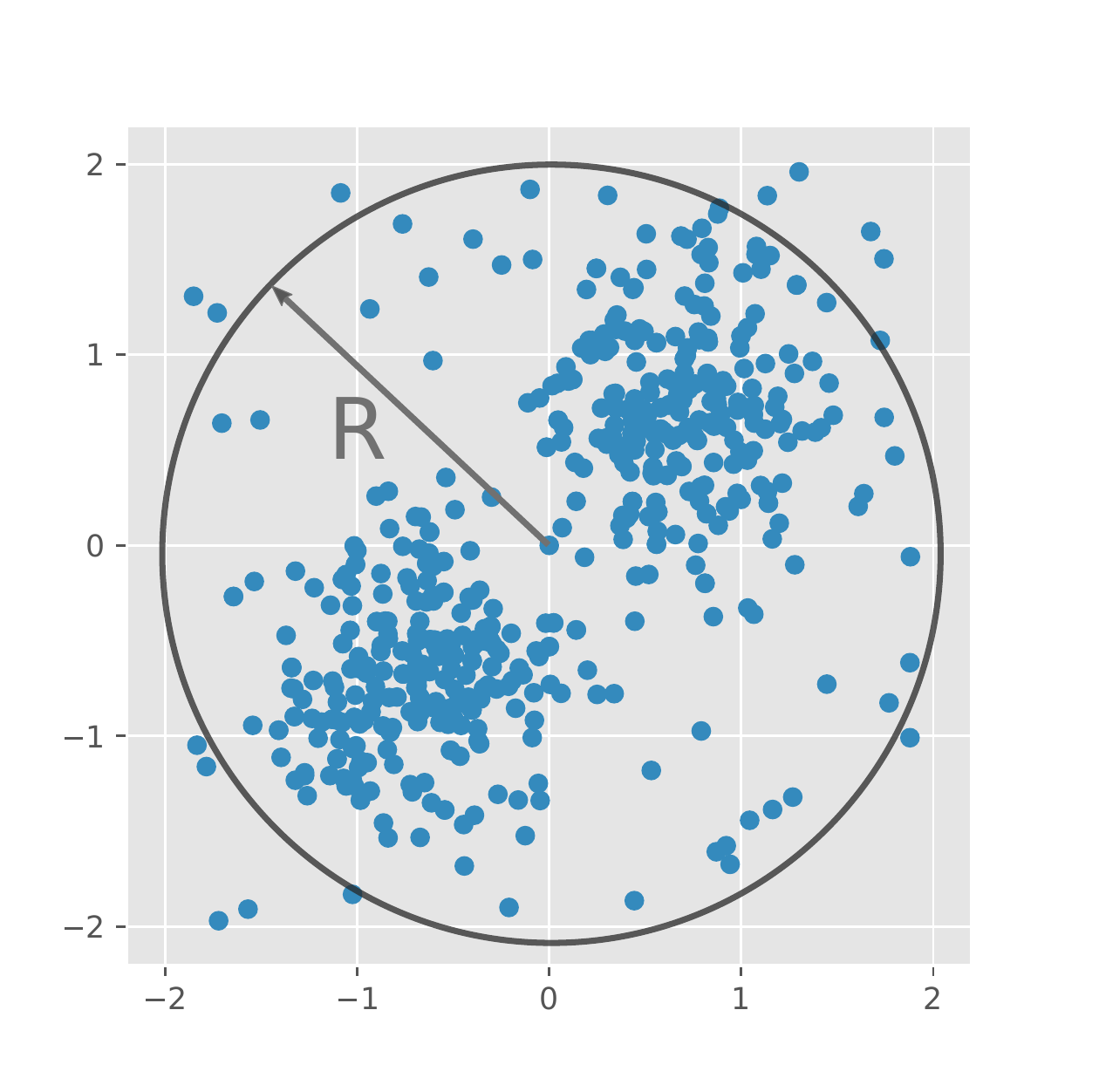}
    \end{subfigure}
    \begin{subfigure}[b]{0.3\textwidth}
        \includegraphics[width=\textwidth]{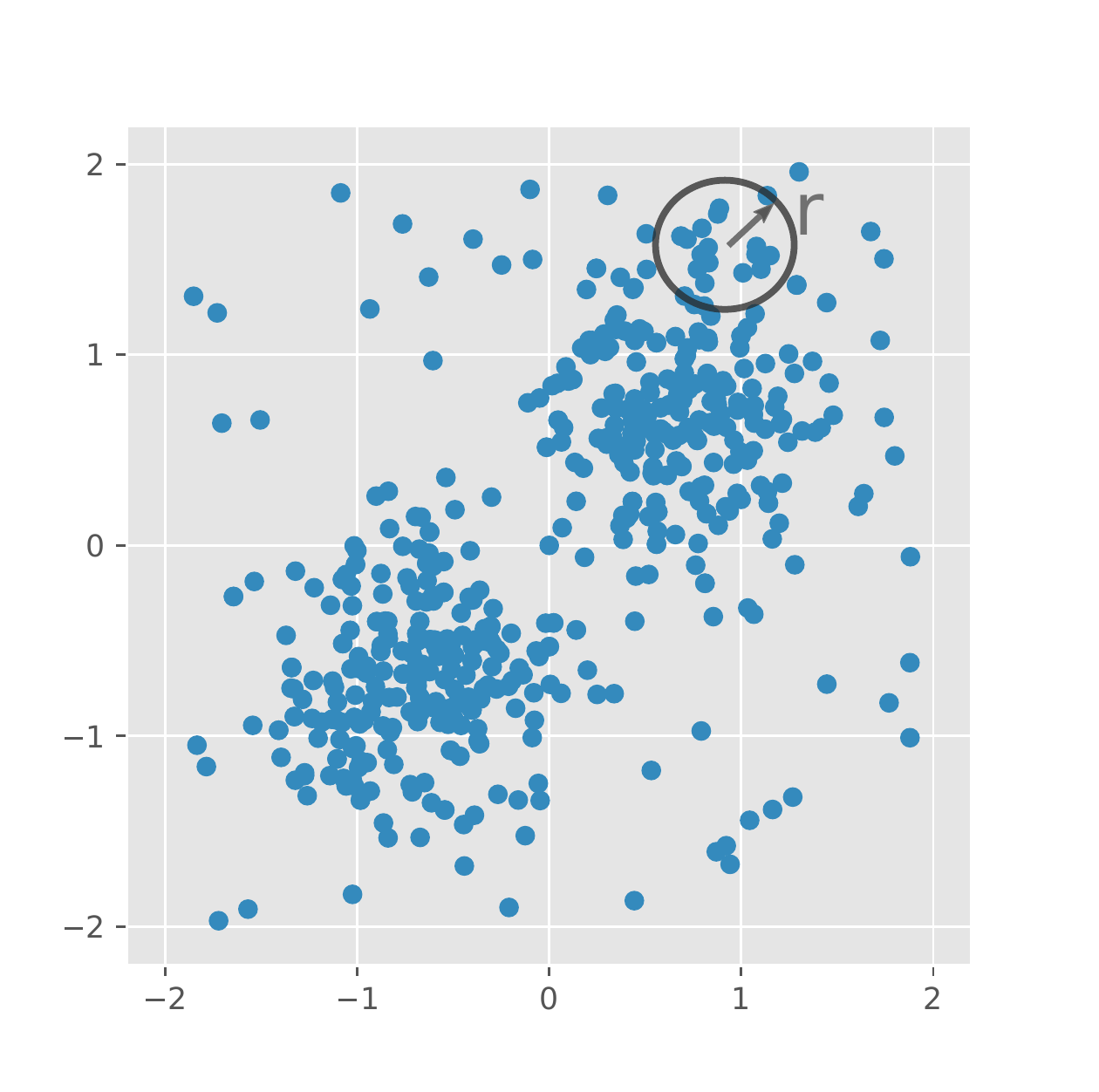}
    \end{subfigure}
        \caption{Two clusters with points random gaussian distributed inside each cluster.}
        \label{fig:pointsuniform}
\end{figure}

Figure~\ref{fig:pointsuniform} illustrates a possible prototype in a 2D feature space, involving two circular clusters of normally distributed data.  
The distribution of the data is shown in Figure~\ref{fig:pointsuniform}(a).  Two main clusters can be identified at a more macroscopic scale.  Actually, these clusters might not coincide with the partition found if some of the aforementioned criteria were used. This partitioning implies a specific number of points for identifying the clusters, as shown in Figure~\ref{fig:pointsuniform}(b). If the desired clusters are to be searched at a much smaller scale, such as the cluster shown in Figure~\ref{fig:pointsuniform}(c), they will likely be ignored. Since the expected size of the clusters might be a piece of important information for defining their partitions, it is a crucial parameter of our methodology.  

As discussed in Section~\ref{subsec:anatomy} an estimation of the number of outliers in the data is also relevant for cluster identification. Outliers in the data that do not belong to specific clusters should not be assigned to any cluster. Despite being an evident requirement, many cluster identification methods, most notably partitional methods, do not follow this principle. Therefore, we also consider the expected number of outliers as a parameter of the method.

The methodology works as follows. Given the desired cluster size $s$ and a number of clusters $k$, the dendrogram is first obtained and then analysed in a bottom-up approach. Clusters are merged until $k$ clusters having at least $s$ elements are identified for the first time. Since the last cluster merge might generate a cluster having size much larger than $s$, it is checked if the last merge should be undone. Let $u$ represent the cluster formed after the most recent merge, and let $q$ and $t$ indicate the two clusters that were joined to defined $u$. Representing as $s_u$, $s_q$ and $s_t$ the respective sizes of clusters $u$, $q$ and $t$, we calculate $s_m=\max\{s_q, s_t\}$. Then, if $(s_u-s)<(s-s_m)$ the last cluster merge is kept. Otherwise, the final clusters will include the largest cluster between $q$ and $t$, and will not include $u$. If $k$ clusters cannot be identified by the method, the first merge where $k-1$ clusters are identified is used. The same happens if $k-1$ clusters cannot be identified, and so on.

This method is illustrated in Figure~\ref{fig:regionssingle}. The points in the first row yielded, by using the single and the Ward linkage method, the dendrograms depicted.  Figures (a) and (b) depict the results from the Single-linkage algorithm while Figures (c) and (d) from the Ward's algorithm. If the red points are to be detected as the resulting cluster, the respective number of points is determined and used for selecting the corresponding subtree in the dendrogram by using the here proposed methodology, marked in red.

In addition to the identified clusters, outliers are also detected and pruned from the dendrogram using a top-down approach. Starting from the top, it is verified if one of the two clusters joined in the current merge has a size smaller than $c$. If that is the case, the small cluster is considered an outlier. The same criterium is applied to the next merges until a merge having both clusters larger than $c$ is identified. The detected outliers are represented as blue points in the first row and as a blue line in the dendrograms shown in the second row of Figure~\ref{fig:regionssingle}.
       
Some hierarchical clustering methods, such as Ward's, try to minimize the variance of the points inside a cluster while maximizing the inter-cluster distance. This approach might be appropriate when the distribution of the points representing the objects in the feature space is bimodal at the selected scale. On the other hand, if the distribution of the data is unimodal, the minimization of the variance and maximization of the inter-cluster distance might lead to the detection of false clusters. Furthermore, outliers are not taken into account by such methods. For instance, the Ward method applied to the points shown in the first row of Figure~\ref{fig:regionssingle}(c) will generate the dendrogram shown in the second row of Figure~\ref{fig:regionssingle}(c), where no outlier points were detected. 

It is important to identify cases when a linkage methodology, and respective strategy for cluster identification, lead to the detection of false clusters. Thus, a criteria for quantifying the quality of the obtained partition, henceforth called \emph{relevance}, needs to be defined. Figure~\ref{fig:regionssingle}(d) shows the result of the application of the Ward-linkage method to a bimodal distribution of points. The nucleus of the green cluster was not detected correctly. Comparing the results depicted in Figure~\ref{fig:regionssingle}, it is clear that the nuclei identified by the single-linkage method in both the unimodal and bimodal distributions are more compact and central than those identified by the Ward method. As a result, the nuclei identified by the single-linkage method were found to be more relevant than those detected by the Ward method.

The above considerations suggest that the relevance of a cluster identified by the methodology can be measured in terms of the height of the red line indicated in the respective dendrograms. Whenever a single cluster is identified, its relevance is given by the dendrogram height where the last merge occurred. If more than one cluster is detected, the relevance is calculated as the average of the heights of the last merges from each cluster.

\section{Inferring the Types of Clusters}
\label{s:types_ident}

The proposed approach for identifying the clusters in dendrograms and assigning respective relevance intrinsically provides subsidies that can be further considered for identifying the type of point distribution type of the respective dataset.  Such information could be valuable not only for a better understanding of the analyzed data, but also be used for tuning and enhancing the cluster detection methodology.  Interestingly, the identification of the type of point distributions represents itself as a pattern recognition approach applied to pattern recognition.

Here, we illustrate the above possibility with respect to identifying the type of point distribution in unimodal datasets.
The first step while trying to identify the type of distribution in a set of points is to define a reasonable and effective set of respective features.  Natural candidates are the position of the outliers height, the cluster height, as well as the size of the outliers set and the size of the detected cluster set.

\begin{figure}[t]
    \centering
        \includegraphics[width=0.9\textwidth]{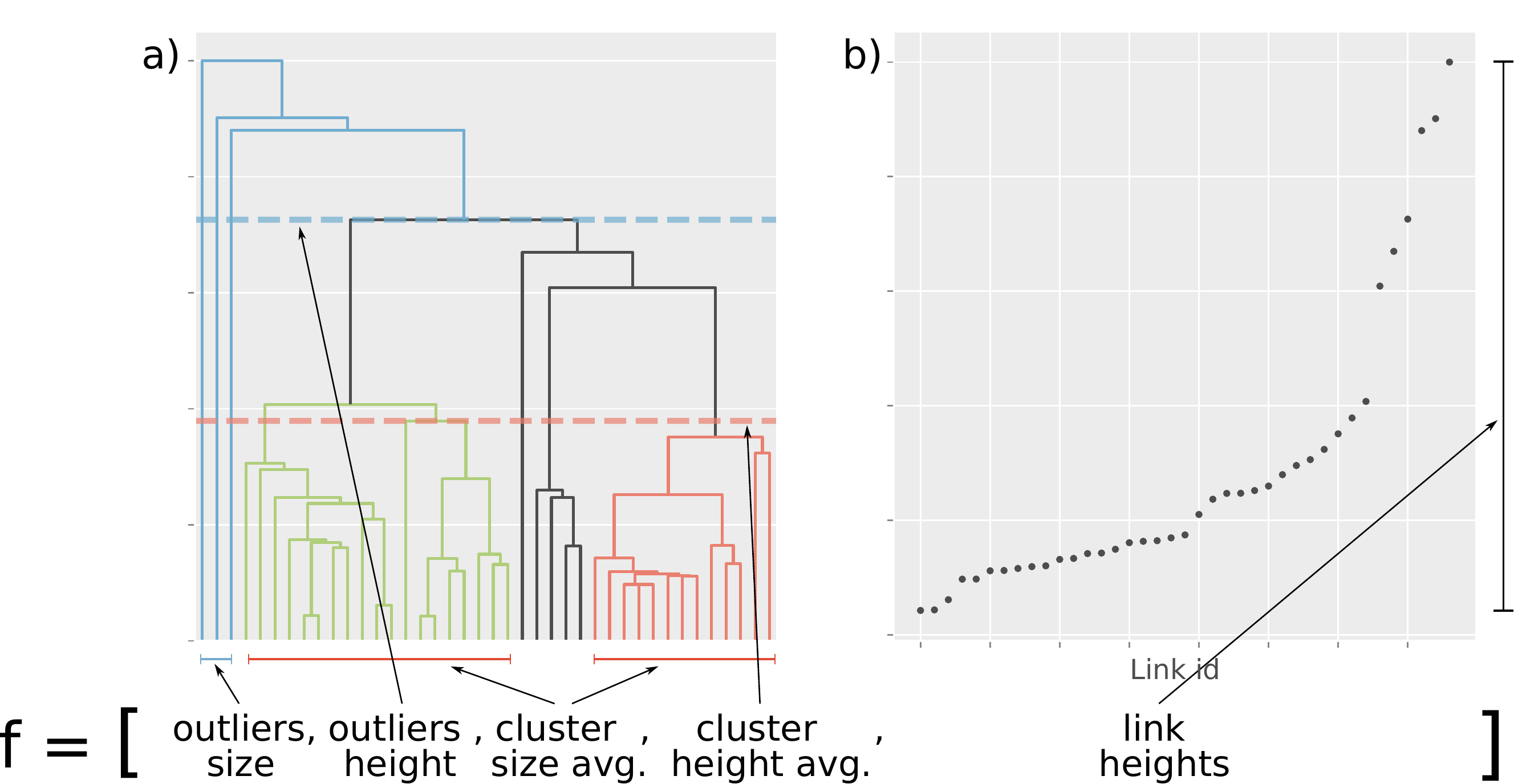}
    \caption{Set of features used for the cluster identification including the dendrogram link heights.}
    \label{fig:featureheights}
\end{figure}

Another potentially useful set of features is related to the number of points that are incorporated into the dendrogram as the respective height variable changes. Figure~\ref{fig:featureheights}(b) depicts an example of the variation of the number of points incorporated into the dendrogram (y-axis) in terms of the heights (x-axis).  One possibility is to consider the whole curve of the number of points $\times$ height.  But since the number of elements in these curves can vary among different datasets, it is necessary to implement an interpolation of the curve, allowing respective resampling with a constant number of elements.

Figure~\ref{fig:featureheights} illustrates the whole feature vector \textbf{f} that can be defined while incorporating the whole set of elements in the number of incorporated points curve.

\begin{figure}[t]
    \centering
        \includegraphics[width=0.9\textwidth]{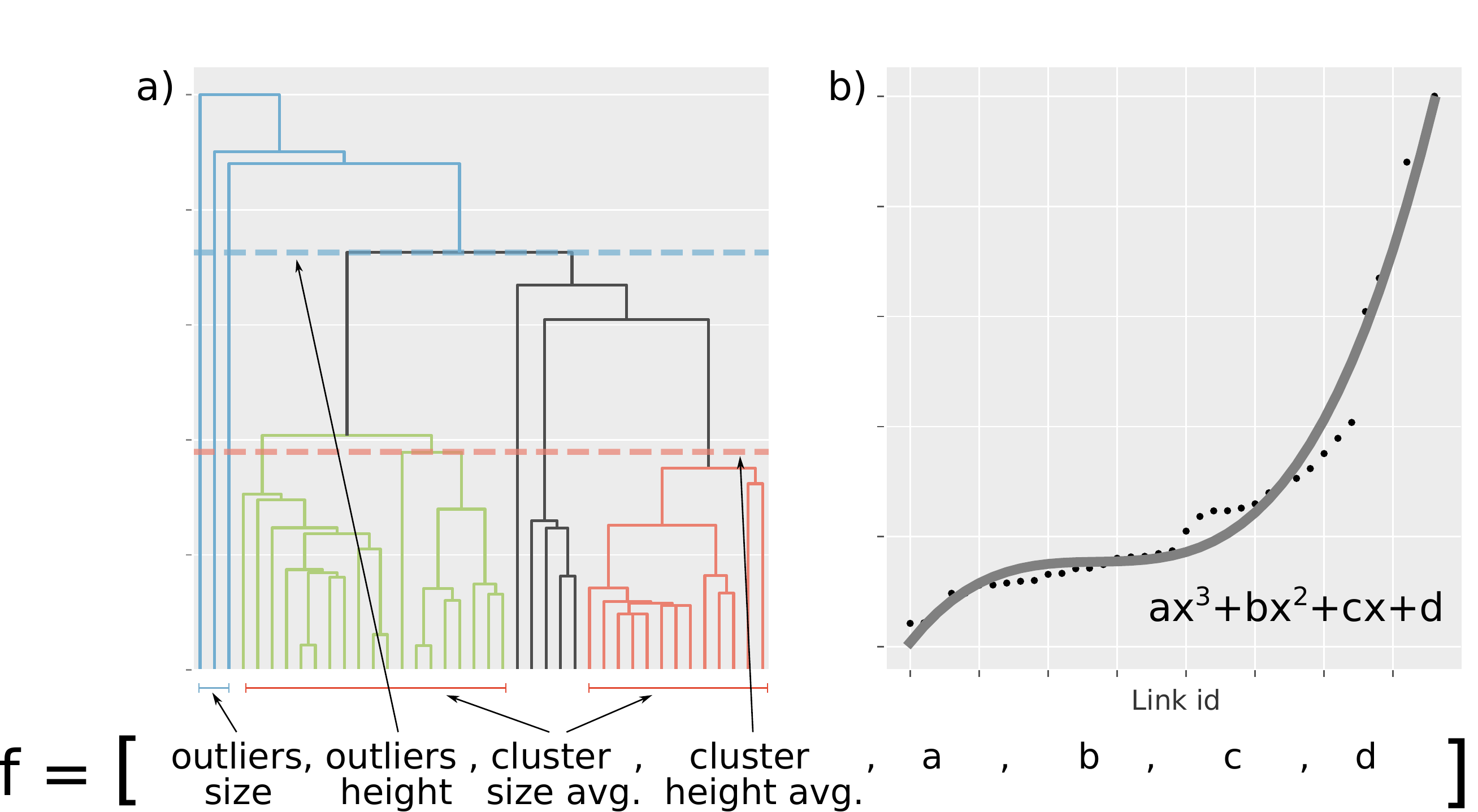}
    \caption{Set of features used for the cluster identification, including the curve fit.}
    \label{fig:featurepolyfit}
\end{figure}

Another possibility is to fit the curve expressing the number of incorporated points by a polynomial. We have found that a cubic
polynomial, i.e~$y = a x^3 + b x^2 + c x + d$, where $x$ is the dendrogram height, tends to provide good fitting for the obtained number of incorporated elements curves.  Figure~\ref{fig:featurepolyfit} shows the alternative feature vector that can be therefore obtained. 

Once feature vectors have been obtained for a representative set of point distributions of diverse types, it is possible to apply principal component analysis (PCA, e.g.~\cite{wold1987principal}) to investigate possible formation of clusters respective to the considered types of point distributions.

\section{Experiments and discussion}
\label{sec:experiments}

Eight synthetic data distributions were considered, including unimodal and bimodal configurations derived from random distributions. The four unimodal distributions were the uniform, Gaussian, power-law and exponential. For each distribution, $D$ independent and identically distributed random variables were used, which defined the dimension of the generated dataset. The bimodal data corresponding to a given zero-centered unimodal distribution was generated using the following approach. Two sets of points $S_1$ and $S_2$ were drawn from the unimodal distribution and the respective standard deviations $\sigma_1$ and $\sigma_2$ of the generated points were calculated. Then, the average $\sigma=(\sigma_1+\sigma_2)/2$ of the standard deviations was calculated, and used for defining a distance $d$ given by

\begin{equation}
d = \alpha \sigma/2
\end{equation}
Next, the coordinates of the points in set $S_1$ were all translated by $d$, while the coordinates of points in $S_2$ were translated by $-d$. A value of $\alpha=4$ was used for all distributions. Thus, $\alpha=2d/\sigma$ is the same for all bimodal distributions. The procedure for generating a power-law bimodal distribution was slightly different. Before translation, the coordinates of the points in set $S_2$ were multiplied by $-1$, so that the tails of the distributions of the two sets of points were placed on opposing directions. %
The uniform distributions were calculated with unitary radius, the power distributions with degree 2, and the gaussian distributions with the identity matrix as the covariance matrix.

\begin{figure}[t]
    \centering
    \begin{subfigure}[b]{0.41\textwidth}
        \includegraphics[width=\textwidth]{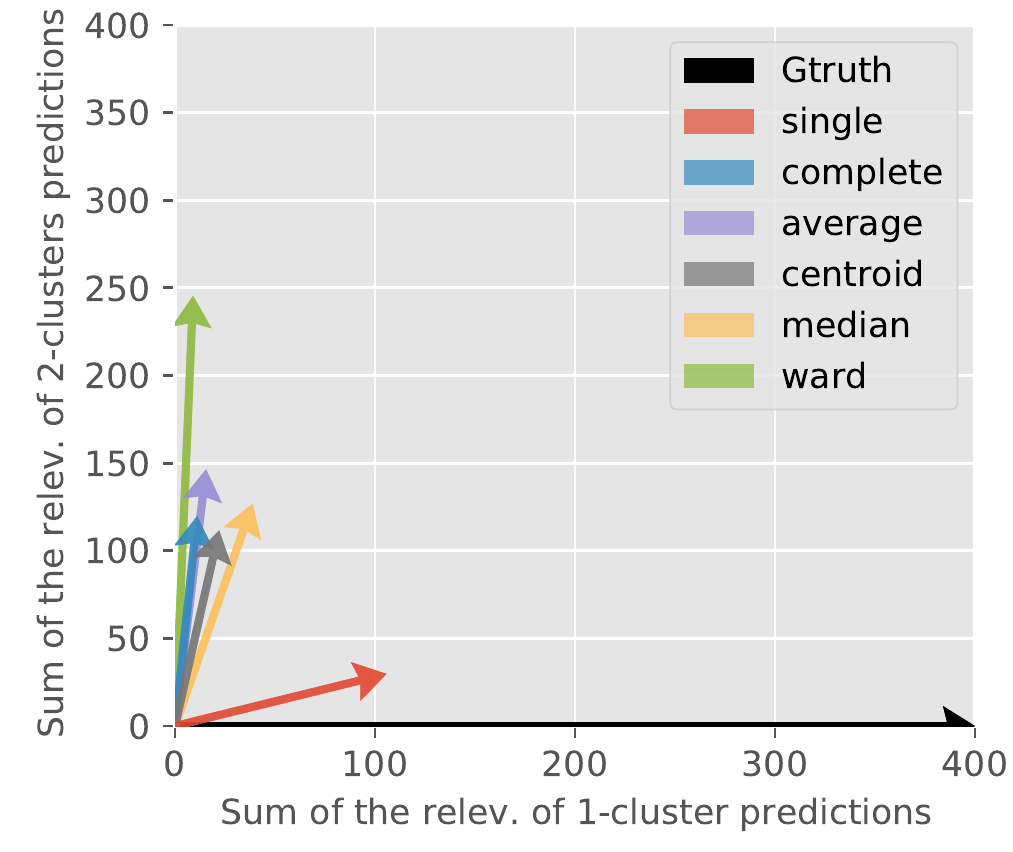}
        \caption{Uniform}
    \end{subfigure}
    \begin{subfigure}[b]{0.41\textwidth}
        \includegraphics[width=\textwidth]{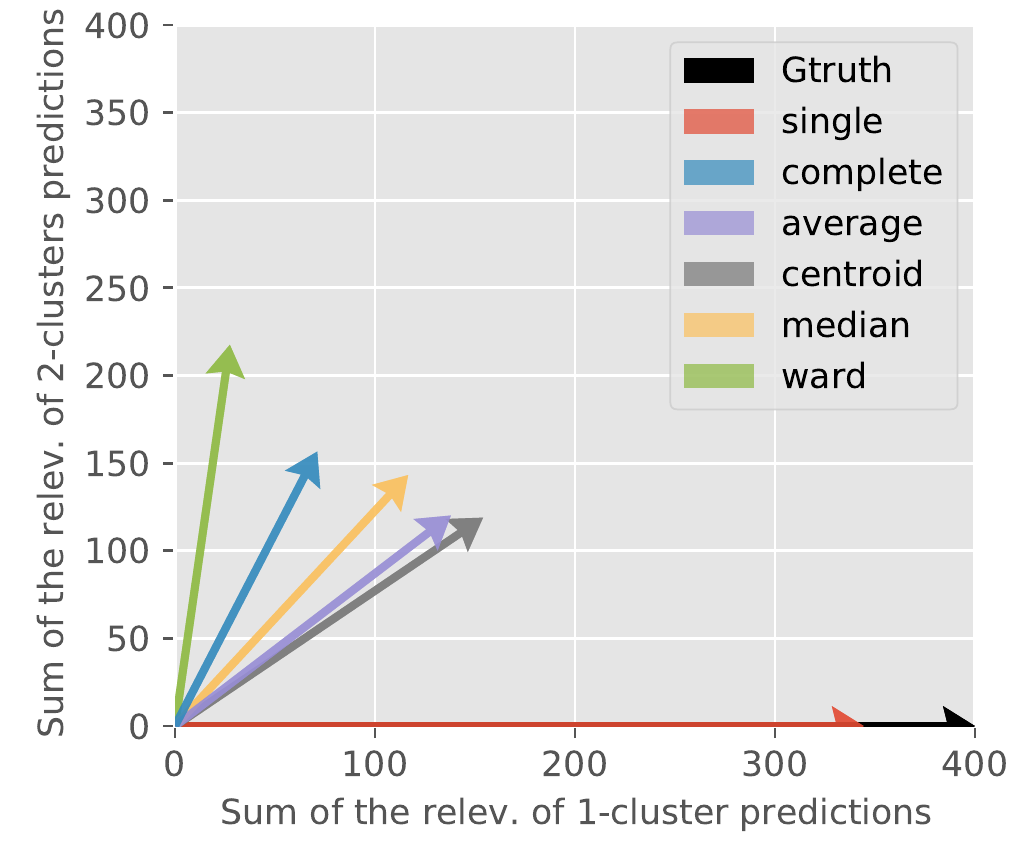}
        \caption{Gaussian}
    \end{subfigure}\\
    \begin{subfigure}[b]{0.41\textwidth}
        \includegraphics[width=\textwidth]{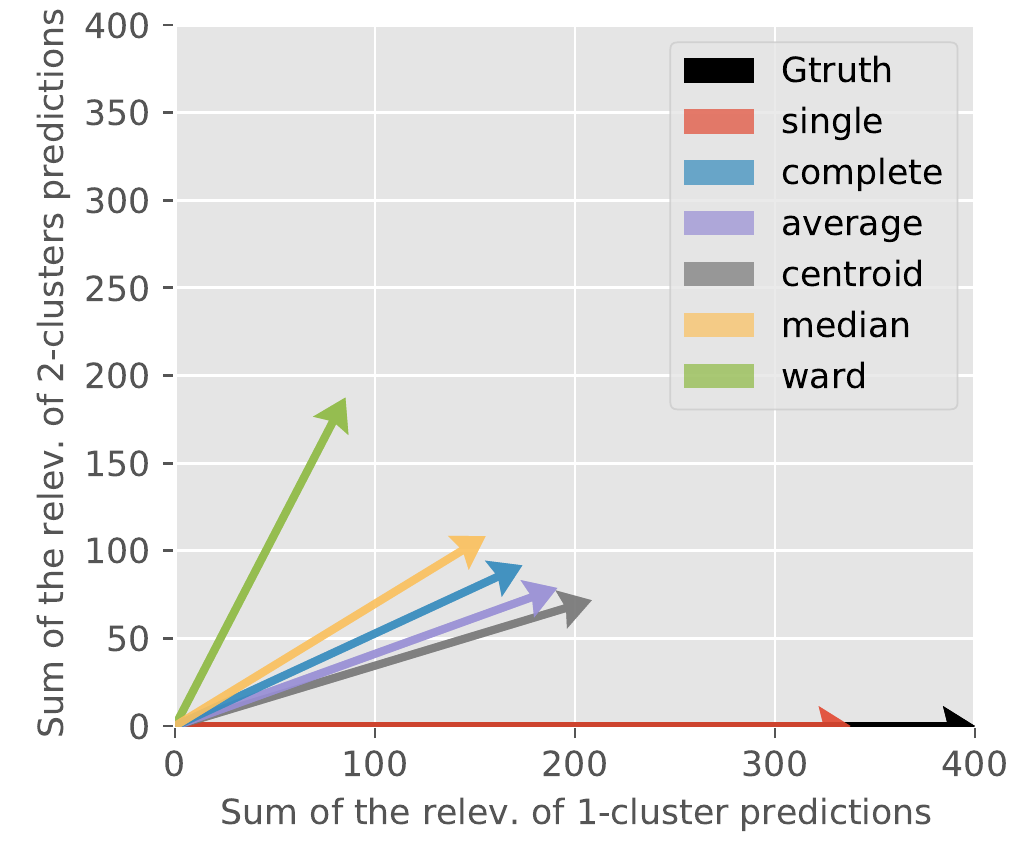}
        \caption{Power}
    \end{subfigure}
    \begin{subfigure}[b]{0.41\textwidth}
        \includegraphics[width=\textwidth]{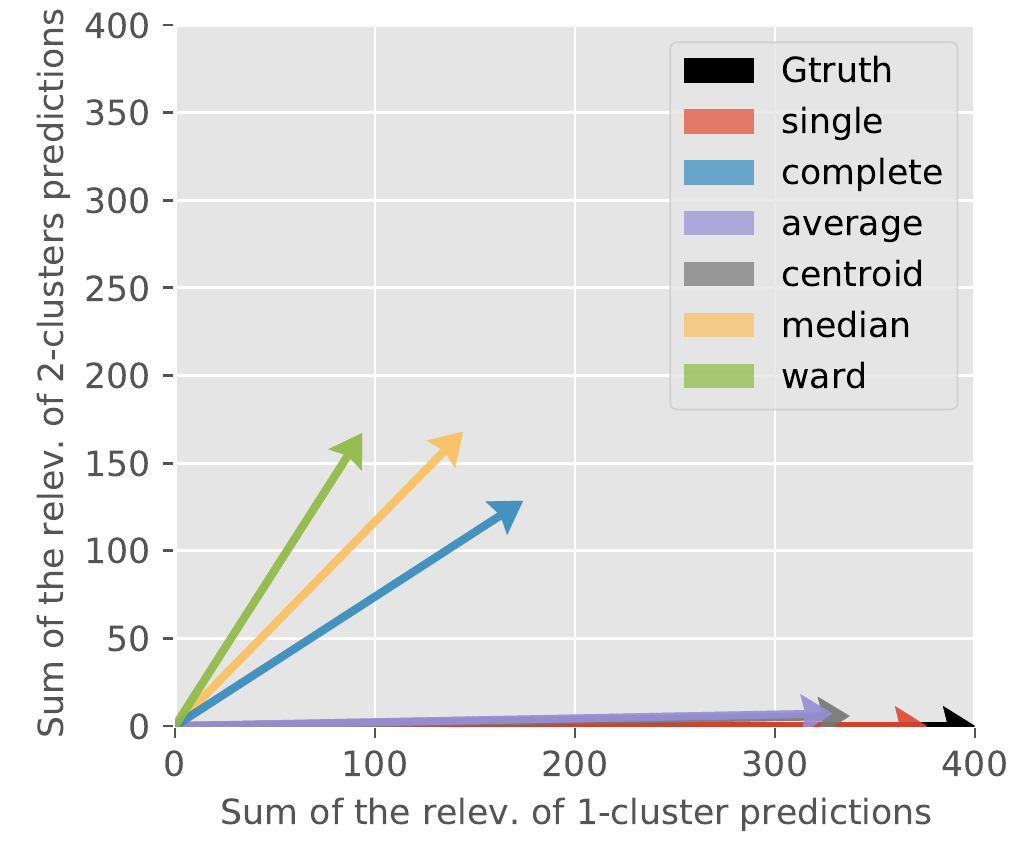}
        \caption{Exponential}
    \end{subfigure}
    \caption{Vectorial visualization of the method proposed evaluated using unimodal distributions.}
    \label{fig:vectorsuni}
\end{figure}

\begin{figure}[t]
    \centering
    \begin{subfigure}[b]{0.41\textwidth}
        \includegraphics[width=\textwidth]{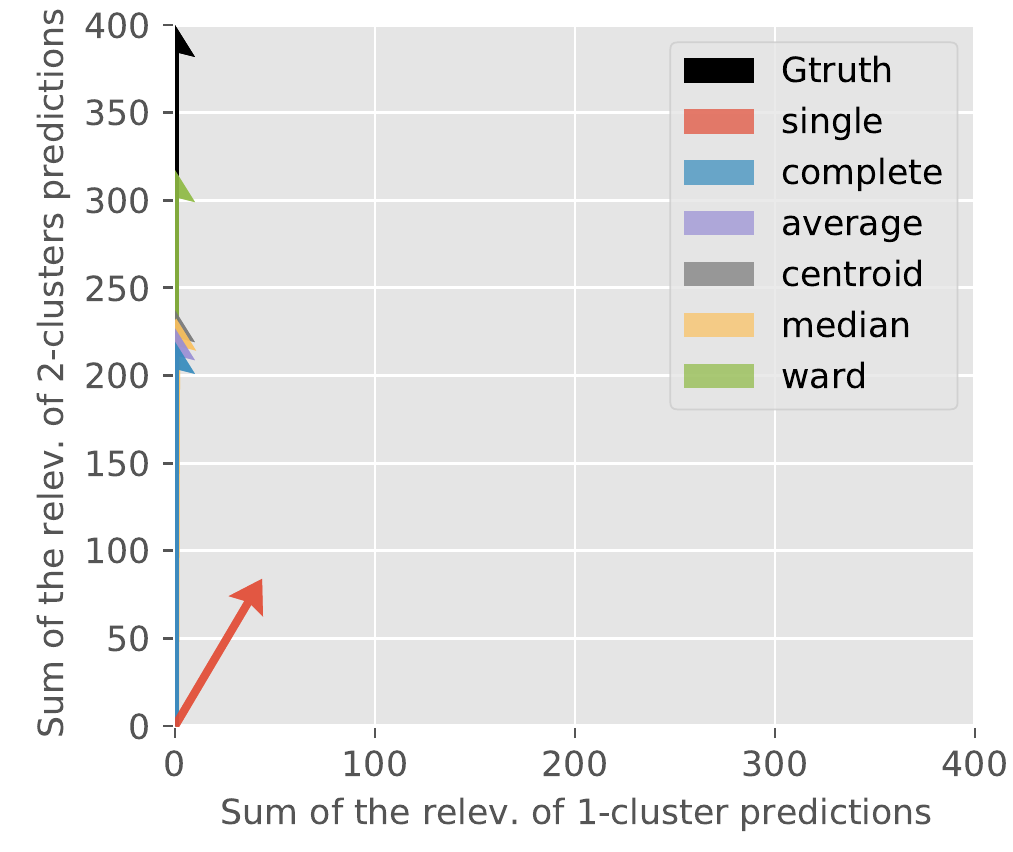}
        \caption{Uniform}
    \end{subfigure}
    \begin{subfigure}[b]{0.41\textwidth}
        \includegraphics[width=\textwidth]{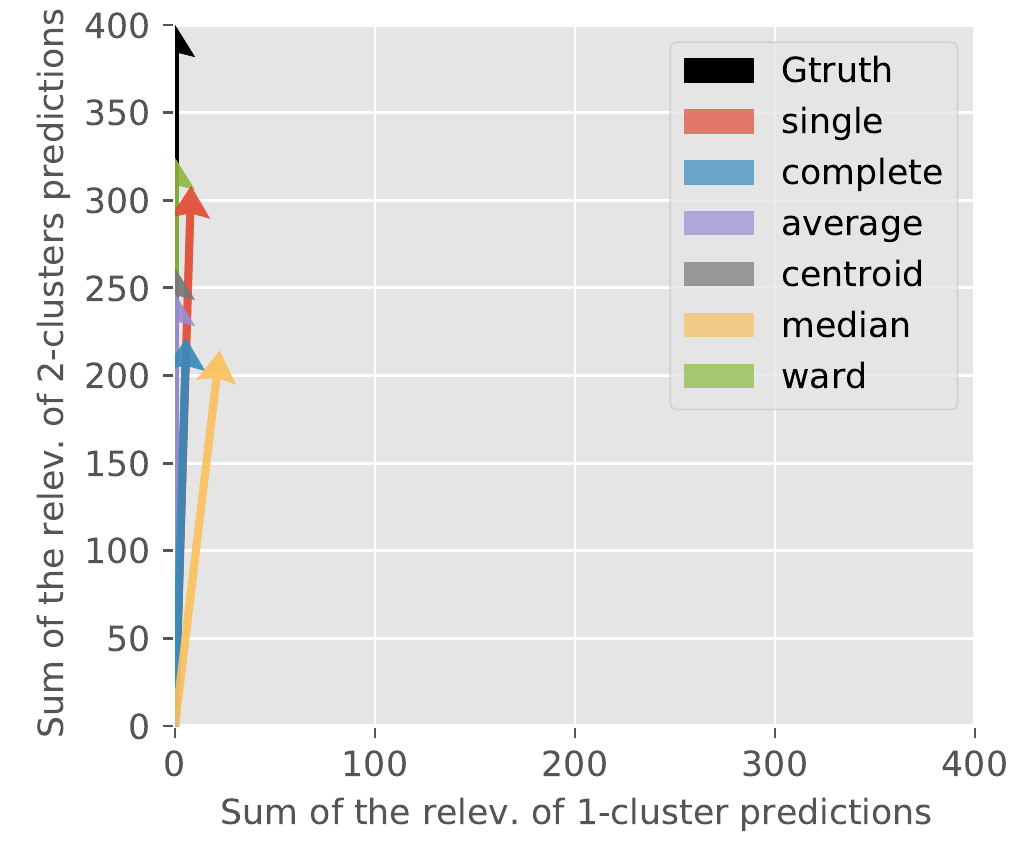}
        \caption{Gaussian}
    \end{subfigure}\\
    \begin{subfigure}[b]{0.41\textwidth}
        \includegraphics[width=\textwidth]{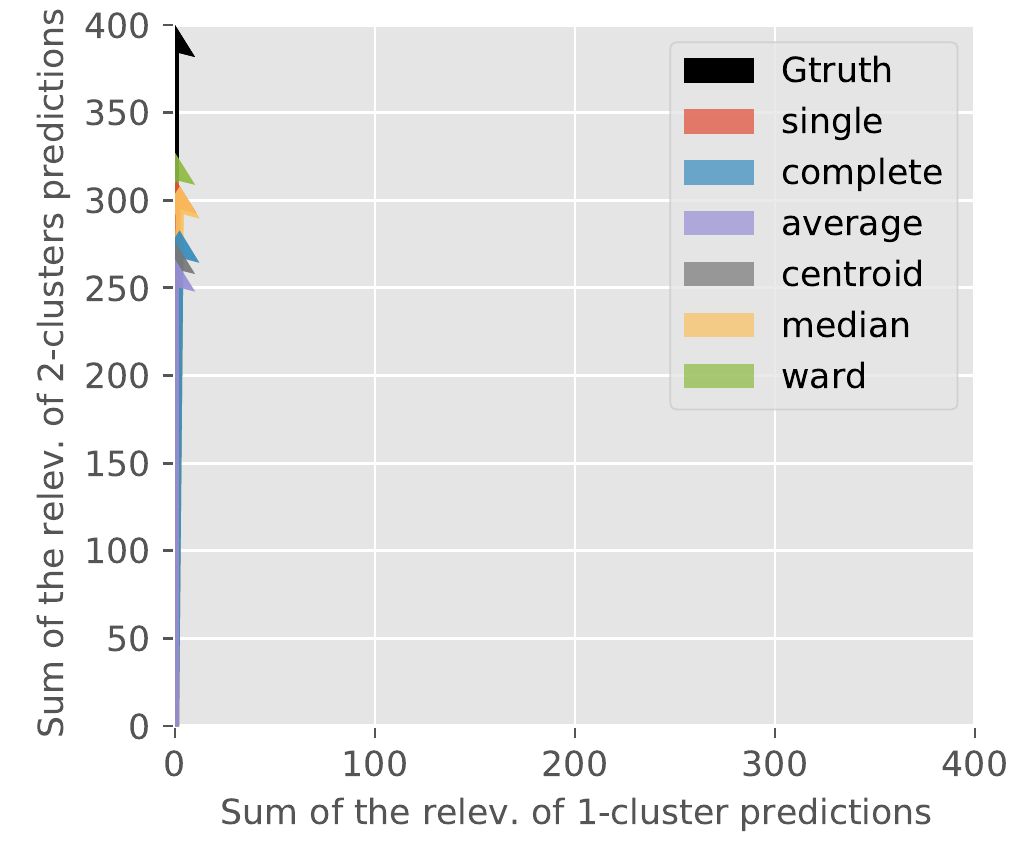}
        \caption{Power}
    \end{subfigure}
    \begin{subfigure}[b]{0.41\textwidth}
        \includegraphics[width=\textwidth]{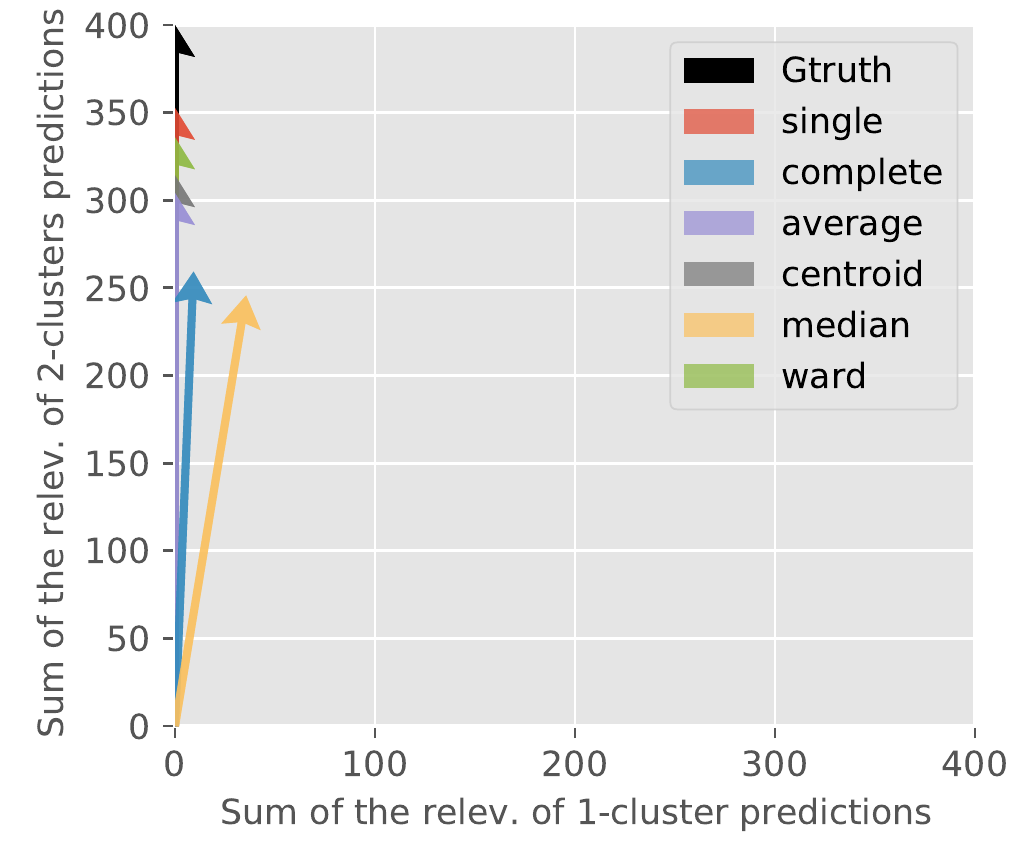}
        \caption{Exponential}
    \end{subfigure}
    \caption{Vectorial visualization of the method proposed evaluated using bimodal distributions.}
    \label{fig:vectorsbi}
\end{figure}

For each considered distribution, $M=400$ datasets containing $N=500$ points were generated. The dimension of the datasets was changed in the experiments, going from $D=2$ up to $D=10$. Thus, for a given number of dimensions $D$, each linkage method was evaluated on $3,200$ datasets. After generating the dendrograms and applying the methodology described in Section~\ref{sec:method} using $K=2$, $S=0.3N$ as cluster size and $P=0.02N$ for removing outliers, the number of detected clusters and the respective relevance of the clusters found were calculated to define the performance vectors described in Section~\ref{sec:method}. The results are shown in Figures~\ref{fig:vectorsuni} and \ref{fig:vectorsbi} for, respectively, unimodal and bimodal data. 

\begin{minipage}[b]{\textwidth}
\begin{minipage}[b]{0.5\textwidth}
        \centering
        \small
        \begin{tabular}[b]{lcc}                                                            
                \toprule
                I & $n_{pred}$. & Relev. \\\midrule
                1 & 2 & \enskip(0.0, 0.89)\\
                2 & 2 & +(0.0, 0.91)\\
                3 & 1 & +(0.6, 0.00)\\
                \midrule
                \multicolumn{2}{l}{Final relev.:}&(0.6, 1.80)\\
        \end{tabular}
        \vspace{1em}
\end{minipage}
\hfill
\begin{minipage}[b]{0.44\textwidth}
        \centering
        \includegraphics[width=\textwidth]{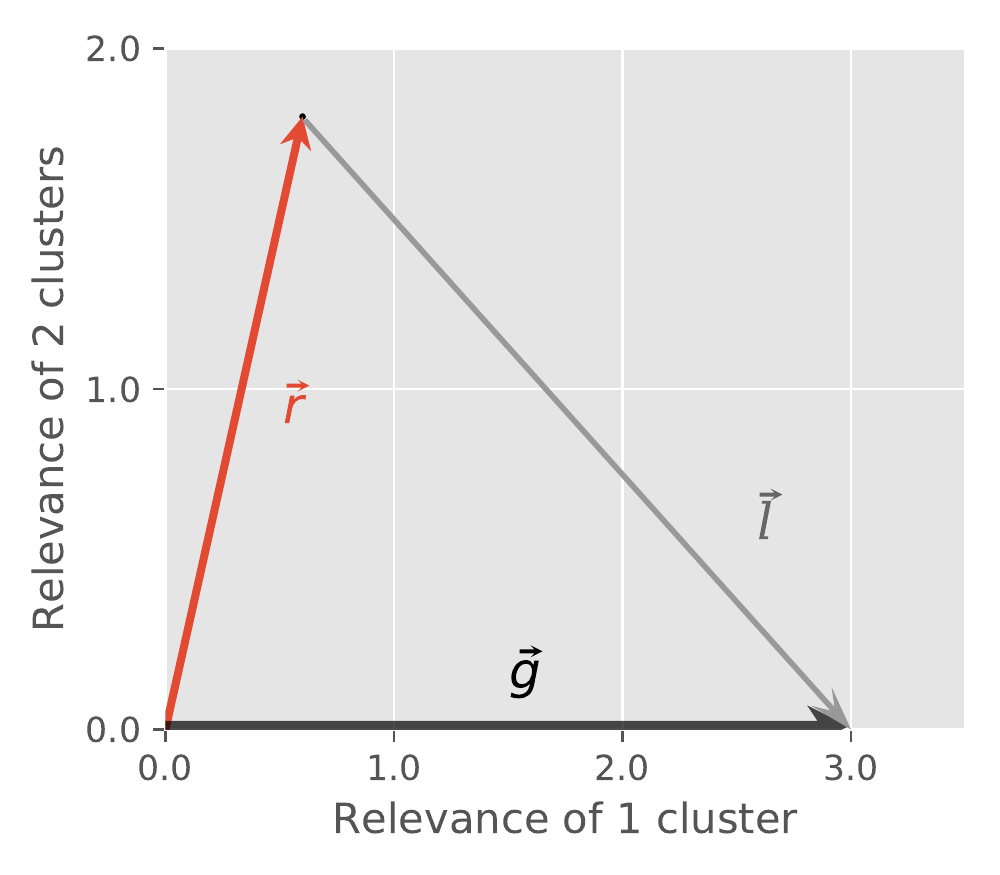}
\end{minipage}
        \captionof{figure}{Example of the computation of the method proposed. Evaluation of the method over three random samples of points (I=1,2,3). Each realization results in a predicted number of clusters ($n_{pred}$) and a corresponding relevance encoded by a tuple, with the first coordinate being the relevance for 1 cluster and the second correspondent to the identification of 2 clusters. The final relevance is obtained by summing the tuples by coordinate.}
        \label{fig:diagramrelev}
\end{minipage}

In all experiments, the hypothesis is that there are two clusters in the data. However, in the case of unimodal data, the linkage methods should yield a dendrogram that when analyzed should not lead to the detection of two clusters. Thus, an ideal method should detect a single cluster with relevance equal to 1 for all the $400$ considered unimodal datasets. This ideal situation is represented as a black vector in Figure~\ref{fig:diagramrelev}. 

The results show that the number of detected clusters varies among the distributions and the linkage methods. For the uniform unimodal distribution, most methods detected 2 clusters in almost all datasets, with the notable exception of the single-linkage method that detected the correct number of clusters in 76\% of the cases. For the gaussian and power-law distributions, the number of detected clusters fluctuates for all linkage methods with the exception of the single-linkage, that detected the correct number of clusters in all cases. The average, centroid, and single linkage methods detected the correct number of clusters in almost all realizations of the exponential distribution.

The reason for the variation in performance among the distributions is the difference between the nuclei of these distributions. While the uniform distribution does not have a well-defined nucleus, the other three distributions contain high-density nuclei that are easier for the methods to detect. This corroborates the importance of considering the nucleus during cluster detection. It is interesting to note that the Ward method resulted in the worst performance in all cases, leading to the detection of false clusters in 82\% of all datasets.

\begin{figure}[t]
    \centering
	\includegraphics[width=0.4\textwidth]{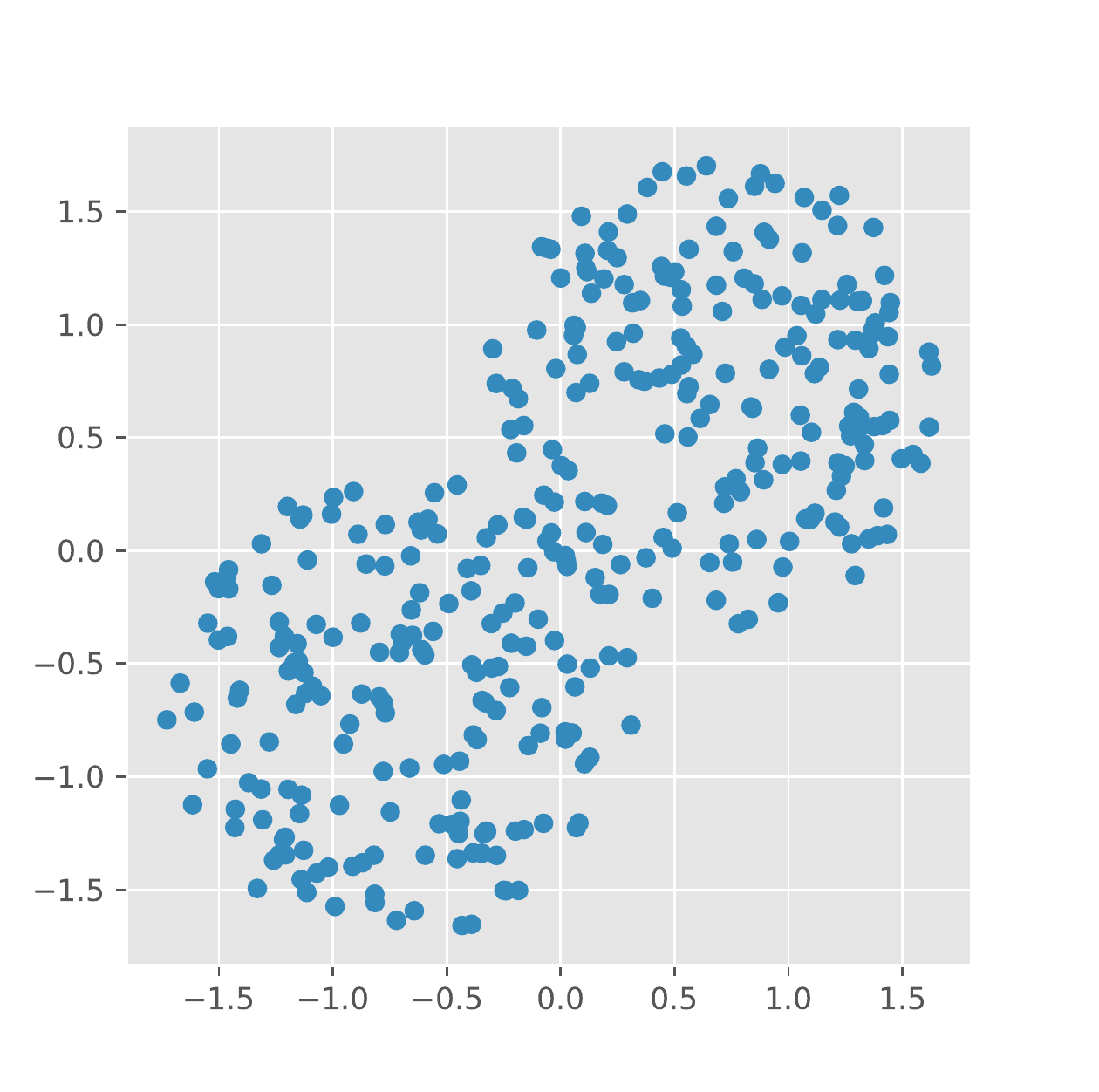}
    \caption{Bimodal uniform distributions of points}
    \label{fig:uniformbi}
\end{figure}

When bimodal data is considered, in almost all situations the methods detected the correct number of clusters, being different only regarding the relevance of the clusters found. The single-linkage method resulted in the lowest relevance for the uniform distribution, that is, it indicates that the two clusters found are not particularly relevant even though it is known that the data have two clusters. This could be interpreted as the single-linkage method not performing well for bimodal data, but we argue that this is not the case for the data used in the experiments. In order to understand why a small relevance should be desired in such situations, Figure~\ref{fig:uniformbi} shows an example of bimodal data generated from the uniform distribution and adopted for evaluating the methods. It is clear that the clusters are very close to one another. Actually, the distances among some points inside the clusters are larger than the distance between the clusters. Thus, depending on the criteria used for defining the clusters, these two clusters could be taken as a single cluster since there is no evident separation between their points. For instance, consider a situation where a clustering algorithm is applied for categorizing apples as ripe and unripe based on color. In this situation, it would be easy to categorize the limiting cases, but a nearly continuous variation would be observed between the two limits. Thus, the intermediate colors would not imply the clustering of the data.

\begin{figure}[t]
    \centering
    \begin{subfigure}[b]{0.24\textwidth}
        \includegraphics[width=\textwidth]{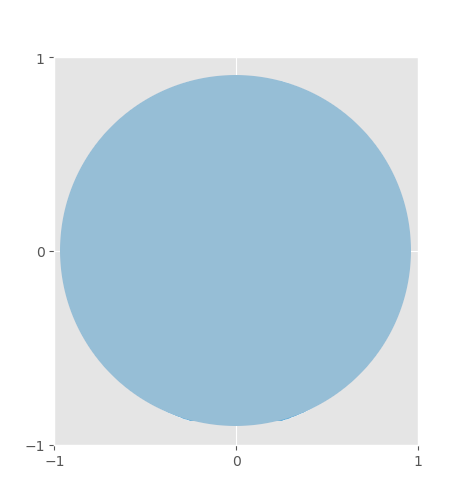}
        \caption{Uniform}
    \end{subfigure}
    \begin{subfigure}[b]{0.24\textwidth}
        \includegraphics[width=\textwidth]{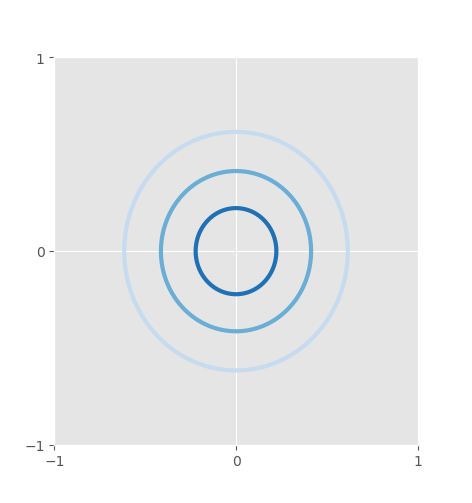}
        \caption{gaussian}
    \end{subfigure}
    \begin{subfigure}[b]{0.24\textwidth}
        \includegraphics[width=\textwidth]{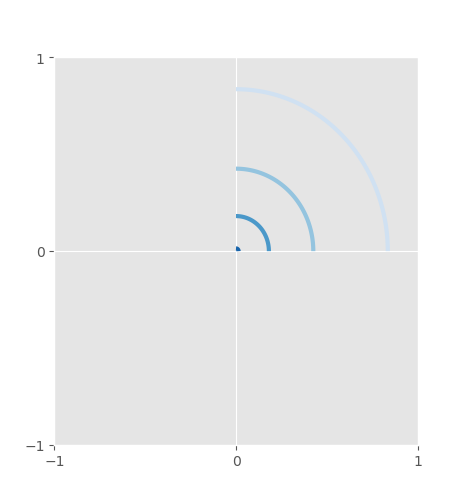}
        \caption{power}
    \end{subfigure}
    \begin{subfigure}[b]{0.24\textwidth}
        \includegraphics[width=\textwidth]{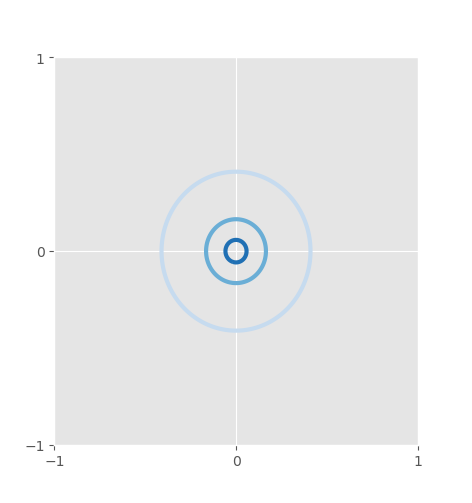}
        \caption{exponential}
    \end{subfigure}
    \caption{Contour plots of the unimodal distributions of points considered in this work. In the first row, five unimodal distributions, from left to right, uniform, linear, power, gaussian and exponential.}
    \label{fig:contourunimodal}
\end{figure}
       
\begin{figure}[t]
    \centering
    \begin{subfigure}[b]{0.24\textwidth}
        \includegraphics[width=\textwidth]{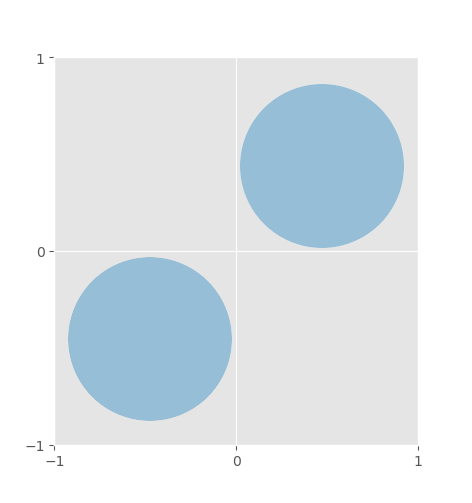}
        \caption{uniform}
    \end{subfigure}
    \begin{subfigure}[b]{0.24\textwidth}
        \includegraphics[width=\textwidth]{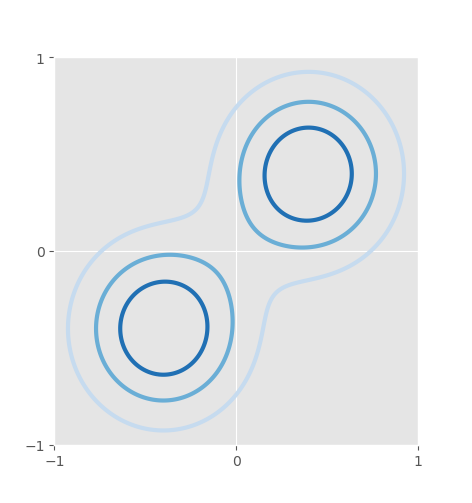}
        \caption{gaussian}
    \end{subfigure}
    \begin{subfigure}[b]{0.24\textwidth}
        \includegraphics[width=\textwidth]{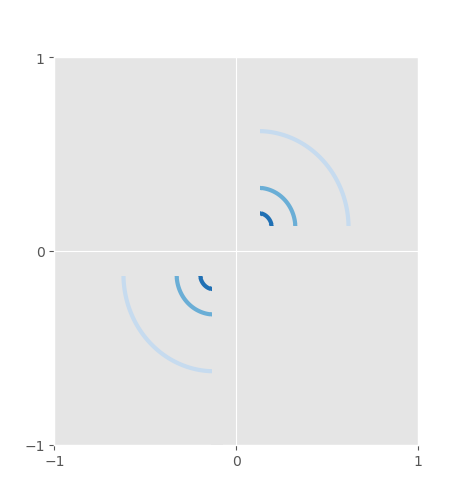}
        \caption{power}
    \end{subfigure}
    \begin{subfigure}[b]{0.24\textwidth}
        \includegraphics[width=\textwidth]{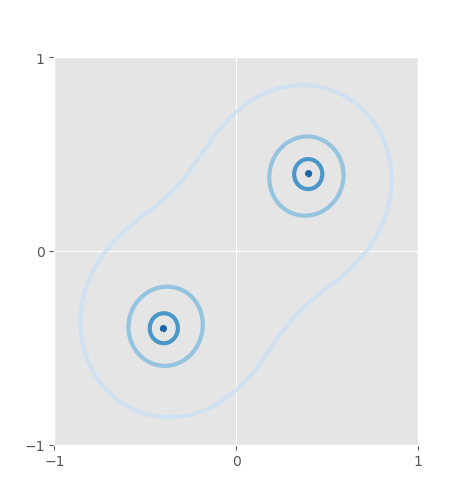}
        \caption{exponential}
    \end{subfigure}
    \caption{Contour plots of the bimodal distributions of points considered in this work. In the first row, five unimodal distributions, from left to right, uniform, linear, power, gaussian and exponential.}
    \label{fig:contourbimodal}
\end{figure}

\begin{figure}[t]
    \centering
        \includegraphics[width=0.97\textwidth]{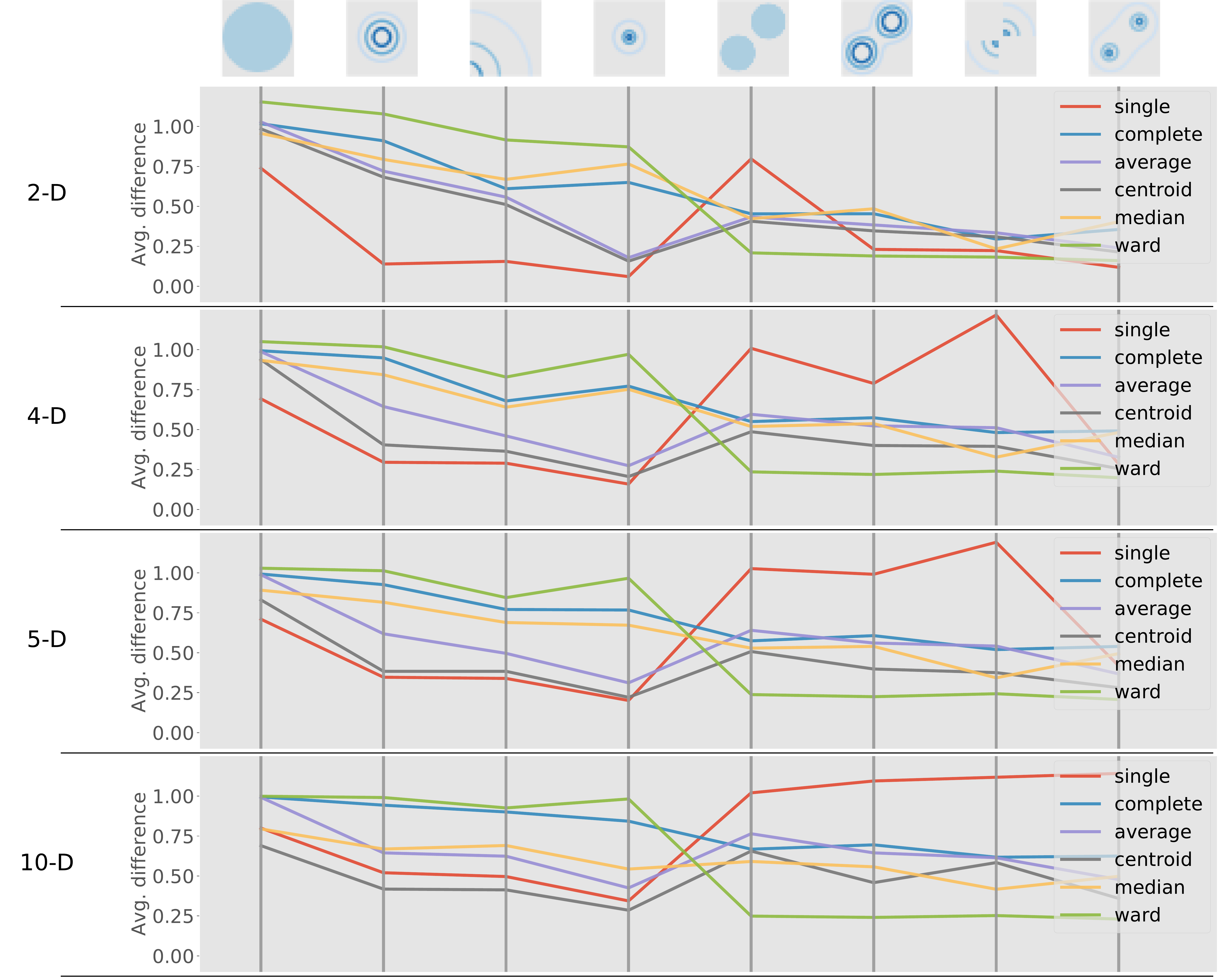}
    \caption{Accumulated error of each method for different data dimensionalities and distributions. Each line represent a linkage method, while each vertical axis represent a data distribution (please refer to Section~\ref{sec:experiments}). }
    \label{fig:accdiffs}
\end{figure}

A summary of the results obtained for each type of dataset is presented in Figure~\ref{fig:accdiffs}. The ordinate axis for each plot indicates the average value of $l$ obtained for a linkage method applied to $400$ datasets having the indicated dimension and distribution. As in the previous results, $k=2$ is used when applying the cluster partition methodology. Figure~\ref{fig:accdiffs} shows that the value of $l$ for the Ward method is very different between the unimodal and bimodal distributions. In the unimodal case, the Ward method tends to detect two clusters with high relevance, as indicated by the results in Figure~\ref{fig:accdiffs}. Therefore, it results in the worst performance among all methods. In the case of bimodal data, it tends to detect two clusters with high relevance, leading to small values of $l$. Also, the performance of the method was very similar for all bimodal distributions, which indicates that the considered distributions did not significantly influence the clusterization. The single-linkage method resulted in the smallest values of $l$ for 2D, 4D and 5D data, which is the desired behavior since in the case of unimodal data $l$ should be small. Interestingly, for 10 dimensions the centroid method performed slightly better than the single linkage. Figure~\ref{fig:accdiffs} also indicates that the complete and median methods tend to have similar performance. The same is observed for the average and centroid methods.
    
To complement our experiments, we also performed an evaluation of the proposed methodology for identifying types of point distributions, described in Section~\ref{sec:experiments}.

\begin{figure}[t]
    \centering
        \includegraphics[width=0.98\textwidth]{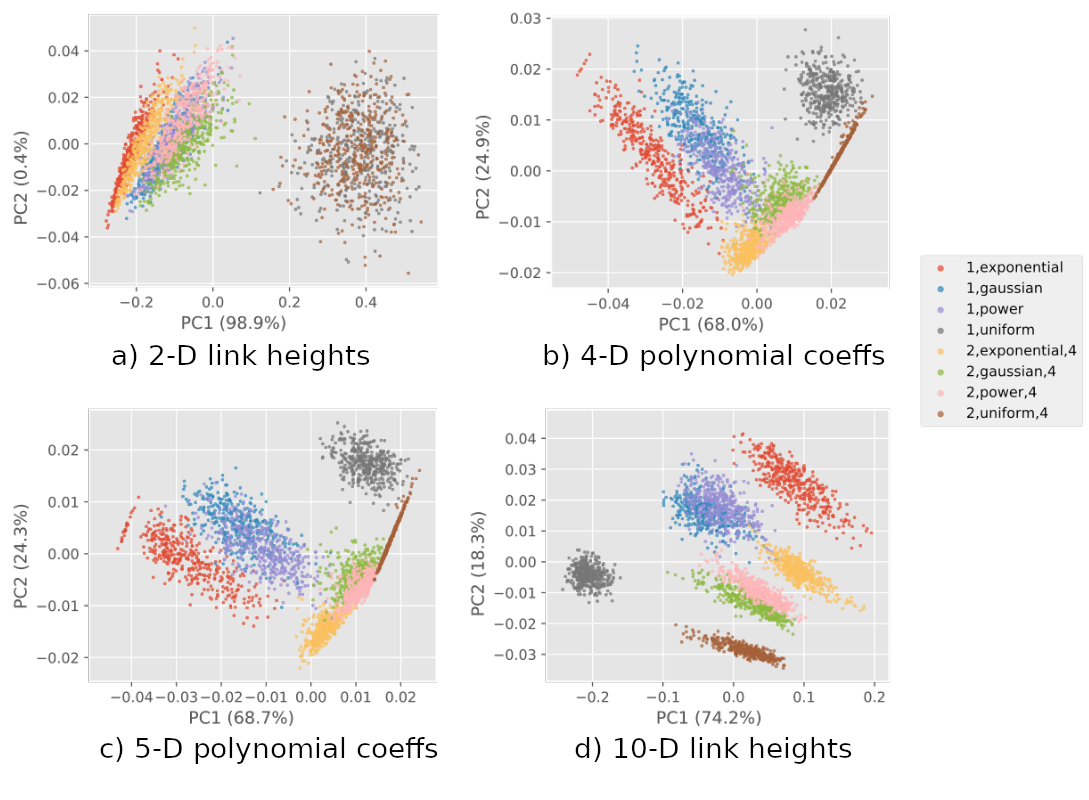}
    \caption{Visualization of the features extracted from the method proposed.}
    \label{fig:featurespca}
\end{figure}

We adopted the same number of realizations, 400, of point distributions previously described. The two alternative feature vectors, i.e.~incorporating all points in the curves of incorporated points as well as the respective fitting by cubic polynomials.  Figure~\ref{fig:featurespca} depicts the results of PCA of the so-obtained feature vectors with respect to point distributions in 2-D (a), 4-D (b), 5-D (c) and 10-D (d). In (a) and (d), the link heights were utilized for the PCA while in (b) and (c) the polynomial fit coefficients were used instead.

Observe that the total variance explanation, shown along the respective principal axes, tends to decrease with the dimension, achieving nearly $100\%$ for 2D.
As can be readily appreciated from the obtained results, several well-separated clusters have been obtained in all considered situations.  As could be expected, the two uniform types of points distributions always appeared further away from the other clusters.  This is a consequence of the fact that uniform distribution of points has no definite nucleus, with all points keeping similar distances one another.   For 2D, except for the two uniform distributions, which completely merged one another while remaining away from the other clusters, the other point distributions tended to occupy respective areas, but with little separation between the groups.  The results obtained for the other 3 dimensions all exhibit well-separated clusters, with a significant separation being observed for 10D.

These results corroborate the potential of the suggested methodology for identifying the types of point distributions in the analyzed data, which could be achieved by applying supervised pattern recognition on the obtained feature spaces.  The identification of the types of clusters has potential not only for better understanding the data under analysis, but also for providing subsidies for enhancing the clustering methodology.  For instance, parameters of the proposed clustering identification method can be fine-tuned adaptively while considering the observed types of clusters.

\section{Conclusion}
\label{sec:conclusion}

The difficulties in comparing clustering methods are, at least partially, related to the definition of what are the properties of the expected clusters.  Actually, this point is also fundamental in devising clustering methods.  In the present work, we aimed at not only of identifying clusters from dendrograms obtained from agglomerative methods, but also assigning respective relevance figures of merit so that the methods could be more objectively compared. More specifically, we defined the relevance of the cluster as the average height of the dendrogram subtrees associated to the detected clusters. Emphasis was placed on quantifying the robustness of the methods to false positives.  We assume that some information is known or hypothesized about the size of the clusters.

A number of interesting results have been obtained and discussed.
Among them, we have that several methods tend to find two clusters
in unimodal data, with the noticeable exception of the single-linkage
method.  At the same time, several methods tended to yield clusters
that do not correspond closely to the nuclei, also including points
from the transition or even outlier zones.  These situations were
duly characterized by lower relevance values. All methods except for the single-linkage were mostly unable to detect outliers, an often desired task.

The obtained results suggest that the single-linkage methodology has some intrinsic potential advantages that motivate this method to be considered in several contexts and applications.  A particularly interesting perspective is to use single linkage as the basis for developing more advanced methodologies, such as the recently proposed hdbscan hybrid approach incorporating distance and density aspects\cite{campello2013density,mcinnes2017accelerated}.  Indeed, the results reported in this work supports single-linkage based methods given their inherent enhanced tolerance to false-positives.  Another related possibility would be to combine agglomerative methods, such as using the single-linkage as a pre-processing step aimed at identifying preliminary cluster candidates to be refined in a subsequent stage, e.g.~Ward's method.

Further interesting prospects can be derived from the described methodology for identification of the types of clusters (Section~\ref{s:types_ident}).  More specifically, given that our results indicate that the type of point distribution can be identified with good accuracy, it would be possible to devise more sophisticated, adaptive, methods in which the types of point distributions are determined as a subsidy for subsequent analysis.   For instance, if a dataset is found to be composed of two types of points distributions, let's say uniform and normal, this information could be used to try to optimize the parameter settings of subsequent cluster identification.  

Other perspectives are allowed by the described results.
For instance, it would be interesting to extend the proposed approach to datasets with larger number of clusters and to other types of point distributions. It would also be interesting to compare divisive 
hierarchical methods.

\section{Acknowledgements}

Eric K. Tokuda thanks FAPESP (grant 2019/01077-3). Cesar H. Comin thanks FAPESP (grant no. 18/09125-4) for financial support.
Luciano da F. Costa thanks CNPq (grant no. 307085/2018-0) for sponsorship.  The authors also aknowledge FAPESP grant 15/22308-2.

\bibliographystyle{unsrt}
\bibliography{main}
\vspace{2em}


\end{document}